\PassOptionsToPackage{table}{xcolor}
\documentclass{article}
\usepackage{arxiv}
\renewcommand{\headeright}{}
\renewcommand{\undertitle}{}

\usepackage{natbib}

\usepackage{amsmath,amsfonts,bm}

\def\eqref#1{equation~\ref{#1}}

\def\1{\bm{1}}

\DeclareMathAlphabet{\mathsfit}{\encodingdefault}{\sfdefault}{m}{sl}
\SetMathAlphabet{\mathsfit}{bold}{\encodingdefault}{\sfdefault}{bx}{n}

\usepackage[dvipsnames]{xcolor}
\definecolor{LinkBlue}{RGB}{0,0,128}
\definecolor{INTACTBlue}{RGB}{235,243,252}
\usepackage{url}
\usepackage{amsmath}
\usepackage{amssymb}
\usepackage{booktabs}
\usepackage{tabularx}
\usepackage{wrapfig}
\usepackage{algorithm}
\usepackage{algpseudocode}
\usepackage{graphicx}
\usepackage{float}
\usepackage{subcaption}
\usepackage{multirow}
\usepackage{placeins}
\usepackage{needspace}
\usepackage{enumitem}
\usepackage{tikz}
\usetikzlibrary{positioning,calc}

\usepackage[colorlinks=true]{hyperref}
\hypersetup{allcolors=LinkBlue}
\newcommand{\method}{INTACT} 
\newcommand{\augsft}{Aug-SFT}   
\newcommand{\nemosft}{Nemo-SFT}  

\title{\method{}: Inducing Latent Reasoning on a Frozen Backbone}
\author{%
  Kutay Tire\thanks{Correspondence to \href{mailto:kutaytire@utexas.edu}{\texttt{kutaytire@utexas.edu}}.} \quad
  Atula Tejaswi \quad Haris Vikalo \quad Sujay Sanghavi \\[2pt]
  The University of Texas at Austin}
\date{}

\begin{document}

\maketitle
\begin{abstract}
Latent reasoning replaces natural-language chains of thought with compact continuous
``soft-token'' representations, reducing the number of intermediate tokens that must be
decoded autoregressively. We use a small external encoder to induce latent
reasoning in a frozen backbone model.
The encoder iteratively maps the backbone's hidden states to new latent blocks,
constructing a latent reasoning prefix, after which the frozen model
writes its remaining reasoning and its answer as text. Our method, \method{}, trains
this encoder through distillation to reproduce the effect of written reasoning steps
while keeping the model's behavior close to that of the frozen backbone, and then
reinforces it on answer correctness. Across four frozen backbones from 1B to 8B,
\method{} improves both GSM8K and average out-of-domain accuracy, including GSM8K gains
from $75.8\%$ to $79.3\%$ on \textsc{Llama-3.2-3B} and from $77.6\%$ to $79.7\%$ on
\textsc{Qwen2.5-3B}. It also reduces end-to-end latency per problem by $19\%$ on
\textsc{Llama-3.2-3B} and outperforms the evaluated latent-reasoning baselines at the 3B
and 8B scales. Beyond the mathematics setting, the same recipe trained on a multi-domain
corpus improves the frozen backbone on twelve evaluation benchmarks spanning mathematics,
science, and commonsense, without task-specific adaptation.
\end{abstract}

\section{Introduction}

%
%
\begin{figure}[H]
\centering
\includegraphics[width=\linewidth]{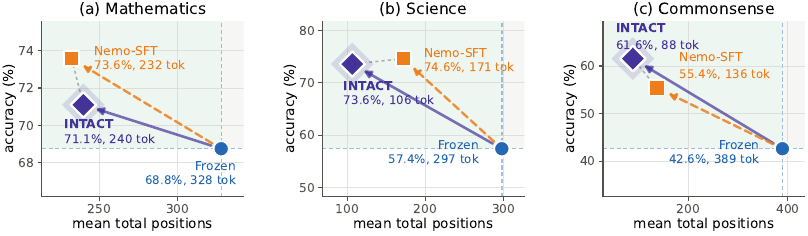}
\vspace{-0.9\baselineskip}
\caption{\textbf{Accuracy and total positions across three reasoning domains.}
A single \method{} encoder is evaluated on mathematics, science and commonsense benchmarks.
\method{} and \nemosft{} are both trained on the
Llama-Nemotron mixture. \method{} trains only the
encoder and leaves \textsc{Llama-3.2-3B} frozen, whereas \nemosft{} updates the language
model via supervised fine-tuning. \hyperref[sec:diversedomain]{Section~\ref*{sec:diversedomain}} describes the
datasets and evaluation setup.}
\label{fig:hero}
\vspace{-0.5\baselineskip}
\end{figure}

Language models are increasingly applied to complex problems across domains such as
mathematics, code, and commonsense reasoning, where the answer follows from a sequence of
dependent steps rather than from recall
\citep{cobbe2021gsm8k,hendrycks2021math,jain2024livecodebench}. In such settings, chain-of-thought prompting draws out the intermediate steps in natural
language and substantially improves accuracy
\citep{wei2022cot,kojima2022zeroshot,wang2023selfconsistency}. That accuracy is paid for in
the output sequence, since every intermediate token must be decoded autoregressively before
the answer appears, and the cost grows with reasoning depth
\citep{chen2024overthinking,snell2024scaling}. 

\emph{Latent reasoning} proposes to move part
of this computation out of language space, replacing some or all of the written steps with
continuous vectors that the model consumes directly \citep{hao2024coconut,shen2025codi}. These vectors may provide compact representations of longer textual traces, and recent work has learned such representations by distilling from
an explicit reasoner \citep{shen2025codi,wei2025simcot}, by multiplexing several reasoning
threads into one sequence \citep{suleymanzade2026mux}, or by looping a transformer over its
own latent state \citep{fan2026lotus,saunshi2025looped}.

Yet most latent-reasoning methods update the backbone alongside the latent
mechanism \citep{hao2024coconut,shen2025codi,fan2026lotus,suleymanzade2026mux}.
On capable backbones, such updates risk eroding existing reasoning abilities,
including those needed for tasks beyond the fine-tuning data \citep{xu2025softcot}.
Freezing the backbone leaves the original model intact and recoverable, but places
the burden on the latent interface to produce inputs the frozen model can use. One
response to this constraint is SoftCoT, which derives soft thoughts from the question with
a small assistant \citep{xu2025softcot}. We instead condition a small encoder
on the backbone's hidden states over the question and preceding latent blocks.
This leads to our central question:
\emph{Can we induce latent reasoning on a frozen backbone, improving both its
accuracy and its efficiency?}

%
\begin{figure}[t]
\centering
\includegraphics[width=\linewidth]{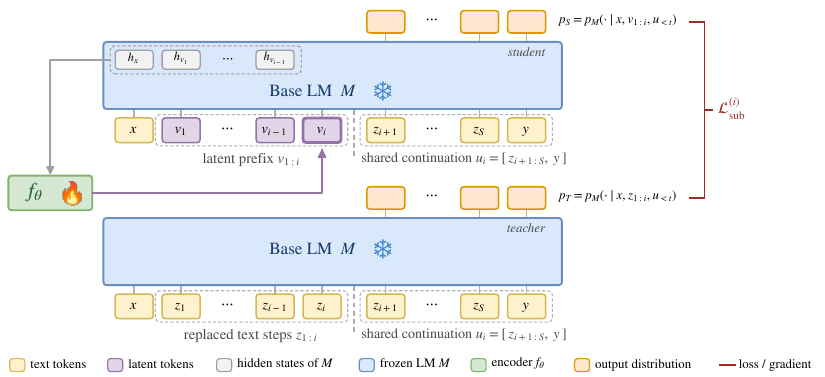}
\caption[Method overview]{\textbf{Stage 1, progressive substitution via
distillation.} The language model $M$ is frozen and only the encoder $f_\theta$ is
trained. At depth $i$, the encoder produces $\mathbf{v}_i$ from $M$'s states over
$\mathbf{x}$ and the preceding blocks $\mathbf{v}_{1:i-1}$, completing the student
prefix $[\mathbf{x},\mathbf{v}_{1:i}]$ against the teacher's
$[\mathbf{x},\mathbf{z}_{1:i}]$. The substitution loss $\mathcal{L}_{\mathrm{sub}}$
distills $p_T$ into $p_S$ over the shared continuation $\mathbf{u}_i$, without directly
aligning $\mathbf{v}_j$ and $\mathbf{z}_j$. \textbf{Stage 2, latent-policy
refinement}, not shown, then refines the same encoder for answer correctness
using group-relative policy gradients, leaving $M$ frozen.}
\label{fig:overview}
\end{figure}

We propose \textbf{\method{}}, illustrated in \hyperref[fig:overview]{Figure~\ref*{fig:overview}}. An
encoder reads the frozen language model's final-layer states and produces $K$ latent
blocks of $c$ vectors each ($K=4$ and $c=4$ in our experiments). During training, the first $j$ blocks learn to replace the first $j$ reasoning steps. Block $j$ is generated from the question representation
together with the frozen model's representation of blocks
$1,\ldots,j-1$. At inference, the resulting $Kc$ vectors are inserted after the question
in place of the corresponding initial textual reasoning, after which the
frozen model generates a solution in its own words.

Training proceeds in two stages.
Stage~1 trains each block to reproduce the effect of the written reasoning step it replaces,
while a preservation term keeps the latent-conditioned model's behavior close to the
frozen model's own. Stage~2 then refines the encoder against answer correctness using group-relative advantages. Across four backbones from 1B to 8B, \method{} improves
upon the frozen backbone while reducing the number of generated tokens, and it surpasses the evaluated latent-reasoning methods at the 3B and 8B scales. We summarize our contributions as follows:
\begin{enumerate}
    \item \textbf{Frozen-backbone latent inference.}
    \method{} constructs a latent prefix sequentially in $K$ blocks. Each block
    is produced from the frozen model's hidden states over the question and
    preceding blocks, thereby conditioning the encoder on the model's
    representation of the prefix constructed so far. Once the prefix is
    complete, the frozen model generates its textual reasoning and answer
    without further encoder calls.

    \item \textbf{Two-stage encoder training.}
    Progressive substitution distills the effect of
    successive written reasoning steps into latent blocks, while on-policy
    preservation constrains their effect on generated continuations. A second
    stage refines the encoder on answer correctness using group-relative
    policy gradients.

    \item \textbf{Accuracy with fewer tokens.}
    Under \emph{mathematics training on GSM8K-Aug}, \method{} improves frozen-model accuracy while
    generating fewer tokens across four Llama and Qwen backbones from 1B to 8B.
    It also outperforms the evaluated latent-reasoning baselines at the 3B and
    8B scales (\hyperref[sec:results]{Section~\ref*{sec:results}}).

    \item \textbf{Multi-domain transfer.}
    Under \emph{multi-domain training on Llama-Nemotron}, \method{} improves
    accuracy and generates fewer tokens across twelve evaluation benchmarks
    without benchmark-specific adaptation, including commonsense, a domain
    absent from training (\hyperref[fig:hero]{Figure~\ref*{fig:hero}},
    \hyperref[sec:diversedomain]{Section~\ref*{sec:diversedomain}}). 
\end{enumerate}

\section{Related Work}
\label{sec:related}

\paragraph{Chain-of-thought and test-time scaling.}
Prompting or training models to produce intermediate steps before answering
improves accuracy on hard problems \citep{wei2022cot,kojima2022zeroshot}, and the
gains grow with the language-space computation spent, whether by aggregating
sampled chains \citep{wang2023selfconsistency}, scaling test-time compute
\citep{snell2024scaling,muennighoff2025s1}, or reinforcement learning that elicits
extended traces \citep{deepseekai2025r1}. The cost of these tokens has in turn motivated work on
overthinking and reasoning efficiency \citep{chen2024overthinking}.

\paragraph{Latent and continuous reasoning.}
A complementary line moves reasoning into non-linguistic computation. Pause and
filler tokens add computation without content
\citep{goyal2024pause,pfau2024filler}, and implicit-CoT methods distill the trace
away entirely \citep{deng2023implicit,deng2024explicit}. Coconut feeds last hidden
states back as input embeddings, unrolling a chain of continuous thoughts
\citep{hao2024coconut}, which may encode superpositions of parallel search paths
\citep{zhu2025superposition}. CODI compresses the chain by self-distillation
\citep{shen2025codi} and SIM-CoT adds step-level supervision \citep{wei2025simcot},
while Soft Thinking replaces sampled tokens with probability-weighted mixtures at
inference \citep{zhang2025softthinking}, although the benefits of this training-free approach have been contested
\citep{wu2025singlethreaded}. Others make continuous chains trainable with
reinforcement learning \citep{butt2025softtokens}, move the reasoning loop inside
the network \citep{saunshi2025looped,geiping2025recurrent}, or pretrain with
continuous concepts \citep{tack2025cocomix}. \citet{zhu2025survey} survey the
area. Many of these methods update the backbone, and most stop the model from
writing its own reasoning.

\paragraph{Compression and frozen-backbone latents.}
Text-space compression shortens prompts and traces while keeping the medium fixed
\citep{jiang2023llmlingua}. Compressed chain-of-thought generates dense
contemplation tokens as trace surrogates \citep{cheng2024ccot}, Token Assorted
mixes discrete latent codes with text \citep{su2025tokenassorted}, and latent
diffusion methods denoise reasoning steps in a learned space that is co-trained
for noise tolerance \citep{kang2025ladir,li2022diffusionlm}. SoftCoT prepends soft
tokens from a small assistant model to a frozen backbone and still generates
language-space reasoning afterwards \citep{xu2025softcot}. 
\hyperref[app:related]{Appendix~\ref*{app:related}} compares \method{} with individual prior methods.

\section{Method}
\label{sec:method}

\paragraph{Notation.}
Let $M$ be a language model with vocabulary $\mathcal{V}$ and input-embedding
dimension $d_M$, and let $\mathcal{V}^\ast$ denote sequences of tokens in
$\mathcal{V}$. For a token string
$\mathbf{x}\in\mathcal{V}^\ast$, let $E(\mathbf{x})$ denote the corresponding sequence
of token embeddings. We consider inputs to $M$ that mix a token sequence $\mathbf{x}$
with a sequence $\mathbf{v}$ of non-token vectors, and we write such inputs in embedding
space, so that the concatenation of the two is $[E(\mathbf{x}),\mathbf{v}]$. We write
$H(\cdot)$ for the final-layer hidden states $M$ produces on a mixed input of this kind,
so $\mathbf{h}=H([E(\mathbf{x}),\mathbf{v}])$ denotes its states over the whole sequence.
Where the embedding map is clear from context we abbreviate $[E(\mathbf{x}),\mathbf{v}]$
as $[\mathbf{x},\mathbf{v}]$, and we write conditioning on such an input as
$p_M(\cdot\mid\mathbf{x},\mathbf{v})$. A training example is a triple $(\mathbf{x},\mathbf{z}_{1:S},\mathbf{y})$, where
$\mathbf{x}$ is a question, $\mathbf{z}_{1:S}=(\mathbf{z}_1,\ldots,\mathbf{z}_S)$ is a
reference reasoning trace of $S$ steps with $\mathbf{z}_i\in\mathcal{V}^{n_i}$, and
$\mathbf{y}$ is the final answer. Without a latent prefix, the unmodified backbone
generates a continuation $\hat{\mathbf{y}}\sim p_M(\cdot\mid\mathbf{x})$ containing its
own reasoning and a final answer $\operatorname{Ans}(\hat{\mathbf{y}})$, requiring an
autoregressive decoding step for every token in it. \hyperref[tab:notation]{Table~\ref*{tab:notation}} collects
the remaining symbols.

\subsection{Latent Encoder $f_\theta$}
The latent encoder $f_\theta$ maps a sequence of hidden states $\mathbf{h}$ to a block
of $c$ continuous vectors. Vectors within a block are generated jointly in a
single encoder pass, whereas blocks are constructed sequentially. To produce
block $j$, the encoder reads the frozen model's final-layer states over the
question and the preceding latent blocks:
\[
\mathbf{v}_j =
f_\theta^{(j)}
\bigl(H([E(\mathbf{x}),\mathbf{v}_{1:j-1}])\bigr)
\in \mathbb{R}^{c\times d_M}.
\]
For $j=1$, the input consists only of the question states,
$H(E(\mathbf{x}))$. After each block is appended, the frozen model processes
the extended prefix, providing the hidden states used to generate the next
block. This recurrence is mediated by the frozen model's representations of
the preceding blocks, rather than by feeding their raw vectors directly
back to the encoder. The encoder therefore builds the latent prefix from
representations supplied by $M$, without learning a separate question
encoder.

\begin{center}
\fcolorbox{black!20}{INTACTBlue}{%
\begin{minipage}{0.955\linewidth}
\fontsize{9}{12}\selectfont
\emph{Example.} For a $60$-token question, the frozen model supplies the states
$H(E(\mathbf{x}))$ to the encoder, which returns the first block
$\mathbf{v}_1=f^{(1)}_\theta(H(E(\mathbf{x})))$ of $c=4$ vectors. These are appended to the
input, so the model reads $[E(\mathbf{x}),\mathbf{v}_1]$ and produces
$H([E(\mathbf{x}),\mathbf{v}_1])$.  The encoder reads
the sequence of $64$ hidden states to produce $\mathbf{v}_2=f^{(2)}_\theta(H([E(\mathbf{x}),\mathbf{v}_1]))$,
and so on:
\[
\text{Prefix positions:}\quad
60 \xrightarrow{+\mathbf{v}_1} 64 \xrightarrow{+\mathbf{v}_2} 68
   \xrightarrow{+\mathbf{v}_3} 72 \xrightarrow{+\mathbf{v}_4} 76.
\]
After $K=4$ blocks the prefix holds $76$ positions, $60$ question tokens and $16$ latent
vectors, and the model generates its reasoning and answer with no further encoder calls.
\end{minipage}}
\end{center}

\paragraph{Encoder architecture.}
The encoder projects the context from $d_M$ to a hidden dimension of $d=1024$. It then applies six Transformer layers, each consisting of
cross-attention from the $c$ queries to the context, self-attention among the
queries, and a feed-forward network, with pre-normalization and residual
connections throughout. A final linear layer projects each output to $d_M$.
Each output vector is normalized and rescaled by a shared learned scalar.
The encoder has $107$M parameters.
All parameters are shared across blocks except for the queries $q_j$.

\paragraph{Inference.}
Given a question $\mathbf{x}$, the encoder is invoked $K$ times, after which $M$
generates its reasoning and answer autoregressively from
$[E(\mathbf{x}),\mathbf{v}_{1:K}]$ without further encoder calls.
\hyperref[alg:inference]{Algorithm~\ref*{alg:inference}} states the procedure.


%
\begin{algorithm}[t]
\caption{\method{} inference. The language model $M$ is frozen and the encoder
$f_\theta$ is fixed.}
\label{alg:inference}
\small
\begin{algorithmic}[1]
\Require question $\mathbf{x}$; frozen LM $M$; trained encoder $f_\theta$;
 $K{=}4$ blocks of $c{=}4$ vectors
\Statex $H(\cdot)$ denotes the final-layer states $M$ produces on an
 embedding-space input, as in \hyperref[sec:method]{Section~\ref*{sec:method}}
\For{$j = 1,\dots,K$}
  \State $\mathbf{v}_j \gets f_\theta^{(j)}
    \bigl(H([E(\mathbf{x}),\mathbf{v}_{1:j-1}])\bigr)$
    \Comment{one block per call, $c$ vectors at a time}
\EndFor
\State $\hat{\mathbf{y}} \sim p_M\bigl(\cdot \mid [E(\mathbf{x}),\mathbf{v}_{1:K}]\bigr)$
  \Comment{no further encoder calls}
\State \Return $\operatorname{Ans}(\hat{\mathbf{y}})$
\end{algorithmic}
\end{algorithm}

\subsection{Training}
\label{sec:training}
Training proceeds in two stages, neither of which updates the language model. 
Every parameter of $M$ remains fixed, and although gradients pass through it, only $\theta$ is
updated, which separates learning the latent interface from relearning the reasoning
behavior of the backbone. Stage~1 trains through distillation to replace written reasoning steps with latent blocks, and Stage~2 then refines the same encoder against answer correctness.
\hyperref[alg:training]{Algorithm~\ref*{alg:training}} summarizes both stages.

\paragraph{Stage 1: progressive substitution.}
\label{sec:stage1}
Stage~1 trains successive latent blocks to reproduce the effect of corresponding
written reasoning steps. At substitution depth $i$, the \emph{teacher
context} contains the first $i$ reference steps,
$C_i^{\mathrm{T}}=[\mathbf{x},\mathbf{z}_{1:i}]$, while the \emph{student context} replaces
them with latent blocks, $C_i^{\mathrm{S}}=[\mathbf{x},\mathbf{v}_{1:i}]$. Both are followed
by the same continuation $\mathbf{u}_i=[\mathbf{z}_{i+1:S},\mathbf{y}]$. For two contexts $A$ and $B$,
we define the continuation-level behavioral divergence
\[
 D(A\Vert B;\mathbf{u})
 =\frac{1}{|\mathbf{u}|}\sum_t \operatorname{KL}\!\left(
 p_M(\cdot\mid A,\mathbf{u}_{<t})\,\Vert\,
 p_M(\cdot\mid B,\mathbf{u}_{<t})\right).
\]
The substitution loss at depth $i$ is
$\smash{\mathcal{L}_{\mathrm{sub}}^{(i)}}=D(\smash{C_i^{\mathrm{T}}}\Vert \smash{C_i^{\mathrm{S}}};\mathbf{u}_i)$, self-distillation
through one frozen model, with the reference steps supplying the behavior to reproduce
and both distributions evaluated under teacher forcing on the shared continuation. The
written steps occupy more positions than the latent blocks that replace them, so we shift
the latent span to keep the continuation at the same positions in both.
\hyperref[app:train]{Appendix~\ref*{app:train}} gives the position assignment and the stages it applies to. We optimize
depths $i=1,\ldots,K$ in order, using the substitution loss only for examples with
$S>i$, and take one encoder update at each depth. After every update, $\mathbf{v}_{1:i}$ is
regenerated with the new parameters
before constructing depth $i+1$. Gradients pass through the preceding blocks
and the fixed language model, but only the shared encoder parameters $\theta$
are updated. 

\paragraph{On-policy preservation.}
Because $\mathcal{L}_{\mathrm{sub}}$ is teacher-forced, it cannot see failures that appear
only when the model generates freely. We therefore sample a rollout under the complete
prefix, $\hat{\mathbf{y}}\sim p_M(\cdot\mid\mathbf{x},\mathbf{v}_{1:K})$, and penalize the intervention's effect
along it with $\mathcal{L}_{\mathrm{pre}}=D(\mathbf{x}\Vert[\mathbf{x},\mathbf{v}_{1:K}];\hat{\mathbf{y}})$.
The sampled continuation is held fixed, and the unsteered model supplies the target
distribution, regularizing the latent-conditioned model on prefixes encountered
during free generation. Preservation applies only at $i=K$, so the Stage~1 loss
of \hyperref[alg:training]{Algorithm~\ref*{alg:training}} is
$\smash{\mathcal{L}_{\mathrm{S1}}^{(i)}}=\smash{\mathcal{L}_{\mathrm{sub}}^{(i)}}
+\mathbf{1}[i=K]\,\beta\mathcal{L}_{\mathrm{pre}}$.

\paragraph{Stage 2: latent-policy refinement.}
\label{sec:stage2}
Stage~1 learns from the behavior induced by reference traces.
Stage~2 instead refines the encoder against answer correctness.
The encoder and frozen model together define
$\pi_\theta(\hat{\mathbf{y}}\mid\mathbf{x})=p_M(\hat{\mathbf{y}}\mid\mathbf{x},\mathbf{v}_{1:K})$ with
the $K$ blocks $\mathbf{v}_{1:K}$ built sequentially by the encoder $f_\theta$,
which alone is updated. For each question, we sample $G$
completions $\hat{\mathbf{y}}^1,\ldots,\hat{\mathbf{y}}^G$ and assign the binary
correctness reward $r^g=R(\hat{\mathbf{y}}^g,\mathbf{y})$, where $R$ indicates
whether the extracted answer matches the reference under the answer-comparison
protocol of \hyperref[app:eval]{Appendix~\ref*{app:eval}}. We
standardize rewards within the group to obtain
$A^g=(r^g-\bar r)/(\sigma_r+\epsilon)$ and minimize the per-sequence-normalized policy loss
\[
    \mathcal{L}_{\mathrm{RL}}
    =-\frac{1}{G}\sum_{g=1}^{G} A^g
      \frac{1}{|\hat{\mathbf{y}}^g|}\sum_t
      \log p_M(\hat y_t^g\mid\mathbf{x},\mathbf{v}_{1:K},\hat{\mathbf{y}}_{<t}^g).
\]
Groups in which every completion receives the same reward have zero advantage
and make no update. The Stage~2 block of \hyperref[alg:training]{Algorithm~\ref*{alg:training}} summarizes this update.

%
\begin{algorithm}[t]
\caption{\method{} training. The language model $M$ is frozen throughout; only $\theta$ is learned.}
\label{alg:training}
\small
\begin{algorithmic}[1]
\Require example $(\mathbf{x},\mathbf{z}_{1:S},\mathbf{y})$; frozen LM $M$; encoder $f_\theta$;
 $K{=}4$ blocks of $c{=}4$ vectors
\Statex $H(\cdot)$ denotes the final-layer states $M$ produces on an
 embedding-space input, as in \hyperref[sec:method]{Section~\ref*{sec:method}}
\Statex $D(A \Vert B; \mathbf{u})\!=\!\frac{1}{|\mathbf{u}|}\sum_t \mathrm{KL}\bigl(
  p_M(\cdot|A,\mathbf{u}_{<t}) \Vert p_M(\cdot|B,\mathbf{u}_{<t})\bigr)$
\Statex \textbf{Stage 1: distill the chain into latents}
\For{minibatch, then $i = 1,\dots,K$}
  \Comment{$\mathcal{L}_{\mathrm{sub}}$ uses examples with $S>i$}
  \State $\mathbf{v}_j \gets f_\theta^{(j)}\bigl(H([E(\mathbf{x}),\mathbf{v}_{1:j-1}])\bigr)$,
    \; $j = 1,\dots,i$
  \State $\mathcal{L} \gets D\bigl([\mathbf{x}, \mathbf{z}_{1:i}] \,\Vert\, [\mathbf{x}, \mathbf{v}_{1:i}] \,;\, [\mathbf{z}_{i+1:S}, \mathbf{y}]\bigr)$
    \Comment{$\mathcal{L}_{\mathrm{sub}}$: $i$ latent blocks replace $i$ text steps}
  \If{$i=K$}
    \State $\hat{\mathbf{y}} \sim p_M(\cdot\mid\mathbf{x},\mathbf{v}_{1:K})$; \quad
      $\mathcal{L} \gets \mathcal{L}+\beta\,D\bigl(\mathbf{x}\Vert[\mathbf{x},\mathbf{v}_{1:K}];\hat{\mathbf{y}}\bigr)$
      \Comment{$\mathcal{L}_{\mathrm{pre}}$: match unsteered $M$}
  \EndIf
  \State $\theta\gets\theta-\eta\nabla_\theta\mathcal{L}$
    \Comment{backprop through fixed $M$; update $\theta$ only}
\EndFor
\Statex \textbf{Stage 2: reinforce with answer reward}
\For{minibatch}
  \State $\mathbf{v}_j \gets f_\theta^{(j)}\bigl(H([E(\mathbf{x}),\mathbf{v}_{1:j-1}])\bigr)$,
    \; $j = 1,\dots,K$; \quad sample
    $\hat{\mathbf{y}}^1,\dots,\hat{\mathbf{y}}^G \sim p_M(\cdot \mid \mathbf{x}, \mathbf{v}_{1:K})$
  \State $r^g \gets R(\hat{\mathbf{y}}^g,\mathbf{y})$, \quad
    $A^g \gets (r^g - \bar{r})/(\sigma_r+\epsilon)$
    \Comment{group-relative advantage}
  \State $\theta \gets \theta + \eta \tfrac{1}{G} \sum_g \tfrac{A^g}{|\hat{\mathbf{y}}^g|} \sum_t
    \nabla_\theta \log p_M\bigl(\hat y^g_t \mid \mathbf{x}, \mathbf{v}_{1:K}, \hat{\mathbf{y}}^g_{<t}\bigr)$
    \Comment{per-sequence mean}
\EndFor
\end{algorithmic}
\end{algorithm}


\section{Experiments}

\subsection{Setup}
\label{sec:setup}

\paragraph{\method{} variants.}
We consider two training settings, on separate corpora and never mixed. Each encoder reads one backbone's
hidden states and is trained for that backbone alone.
\begin{itemize}[leftmargin=1.6em,itemsep=0.15em,topsep=0.35em]
  \item \textbf{Mathematics training on GSM8K-Aug} follows prior latent-reasoning work and
        uses $385$K examples \citep{deng2023implicit} whose solutions are
        sequences of compact arithmetic steps. It is trained separately for each of the
        four backbones and evaluated in domain on the GSM8K test set
        \citep{cobbe2021gsm8k} and out of domain on SVAMP, GSM-Hard and MultiArith, over
        which we report an unweighted macro-average.
  \item \textbf{Multi-domain training on Llama-Nemotron} uses the Llama-Nemotron
        post-training corpus \citep{bercovich2025llamanemotron}, approximately $64\%$
        mathematics, $21\%$ science and $14\%$ code. It is trained once, on
        \textsc{Llama-3.2-3B-Instruct}, and evaluated on twelve held-out mathematics,
        science and commonsense benchmarks.
\end{itemize}
The four backbones are \textsc{Llama-3.2-1B-Instruct}, \textsc{Llama-3.2-3B-Instruct},
\textsc{Llama-3.1-8B-Instruct} \citep{grattafiori2024llama}, and the base
\textsc{Qwen2.5-3B} \citep{qwen2024qwen25}. \method{} uses $K=4$ blocks of $c=4$ vectors
in both settings, with the same encoder architecture and objectives throughout and the
language model frozen. \hyperref[app:domainsetup]{Appendix~\ref*{app:domainsetup}} gives the diverse-domain corpus selection
and segmentation into steps, and \hyperref[app:datasets]{Appendix~\ref*{app:datasets}} the twelve evaluation benchmarks.

\paragraph{Baselines.}
We compare against the unchanged \textbf{Frozen LM}, and
against three supervised baselines that update the language model and so measure what
the same supervision achieves when changing the backbone is allowed. \textbf{No-CoT}
trains on the answer alone. \textbf{\augsft{}} trains on GSM8K-Aug, whose steps are
compact symbolic traces, primarily calculator expressions. \textbf{\nemosft{}} instead trains on
the Llama-Nemotron mixture, whose steps are complete natural-language worked solutions. We further compare
with \textbf{CODI}
\citep{shen2025codi}, \textbf{LOTUS} \citep{fan2026lotus}, and \textbf{MUX}
\citep{suleymanzade2026mux}, which
replace written reasoning with continuous or recurrent computation but likewise
update the language model. \hyperref[app:baselines]{Appendix~\ref*{app:baselines}} describes how each baseline was trained or,
where a result is quoted, which table it is taken from.

\paragraph{Evaluation.}
We evaluate on full test splits with generation caps of $768$ tokens for
\textbf{mathematics} and $2{,}048$ for the \textbf{multi-domain} setting, whose
natural-language solutions are longer. We report answer accuracy and mean generated
tokens. Wherever total token use is shown, \method{} is also charged for its $16$ latent
positions. Accuracy for \method{} is averaged over three encoder seeds, with $\pm$ giving the
sample standard deviation where it is shown. Token counts are measured on the single seed
that initializes Stage~2, so the two stages stay comparable. Ablations and controls use
one encoder seed. Quoted baselines retain their source protocols. \hyperref[app:eval]{Appendix~\ref*{app:eval}} specifies the prompts, decoding parameters,
and answer extraction.

\subsection{Main Results}
\label{sec:results}

%
\begin{wraptable}{r}{0.54\textwidth}
\vspace{-1.1\baselineskip}
\centering
\captionsetup{skip=0pt}
\caption{Accuracy (\%) at a 768-token generation cap across four backbones. Both \method{} stages are trained on GSM8K-Aug. \emph{OOD} is the average on SVAMP, GSM-Hard and MultiArith.}
\label{tab:budget768}
\footnotesize
\setlength{\tabcolsep}{3pt}
\renewcommand{\arraystretch}{1.15}
\resizebox{\ifdim\width>\linewidth\linewidth\else\width\fi}{!}{%
\begin{tabular}{@{}l cc cc cc@{}}
\toprule
& \multicolumn{2}{c}{\textbf{Frozen}} & \multicolumn{2}{c}{\textbf{\method{} Stage 1}}
& \multicolumn{2}{c}{\textbf{\method{} Stage 2}} \\
\cmidrule(lr){2-3}\cmidrule(lr){4-5}\cmidrule(lr){6-7}
& \textbf{GSM8K}$\uparrow$ & \textbf{OOD}$\uparrow$ & \textbf{GSM8K}$\uparrow$ & \textbf{OOD}$\uparrow$
& \textbf{GSM8K}$\uparrow$ & \textbf{OOD}$\uparrow$ \\
\midrule
Llama-3.2-1B & 44.0 & 47.8 & 44.2 & 51.3 & \textbf{48.7} & \textbf{54.5} \\
Llama-3.2-3B & 75.8 & 66.5 & 77.9 & \textbf{69.1} & \textbf{79.3} & 68.8 \\
Llama-3.1-8B & 72.3 & 53.9 & \textbf{83.7} & 71.1 & 83.6 & \textbf{72.1} \\
Qwen2.5-3B & 77.6 & 74.6 & 78.0 & 75.0 & \textbf{79.7} & \textbf{75.1} \\
\bottomrule
\end{tabular}%
}%
\vspace{-0.9\baselineskip}
\end{wraptable}

\paragraph{Against the frozen backbone.}
\hyperref[tab:budget768]{Table~\ref*{tab:budget768}} reports accuracy at the 768-token cap, where both stages
improve GSM8K and out-of-domain accuracy over the frozen backbone on all four models.
Stage~2 further improves six of the eight Stage~1 scores. It is $0.3$ points lower on 3B out-of-domain accuracy
and $0.1$ points lower on 8B GSM8K. The largest Stage~1 gains occur at 8B,
where the bare-prompt frozen model often continues generating after producing its answer.
To separate this formatting issue from reasoning accuracy, we also evaluate the frozen
model with its native chat template. This restores reliable termination and raises its
accuracy to $78.4$, but leaves it below both stages of \method{}.
\hyperref[tab:frozen-prompt]{Table~\ref*{tab:frozen-prompt}} reports the comparison.
The size of the gain varies widely across backbones. Relative to the frozen backbone, Stage~2 adds $11.3$ points on GSM8K at 8B and $2.1$ on
\textsc{Qwen2.5-3B}, the strongest frozen baseline in the table. At 1B the Stage~1
encoder is flat on GSM8K ($44.2$ against $44.0$) while gaining $3.5$ out of domain.
\hyperref[tab:budget]{Appendix Table~\ref*{tab:budget}} reports the complete budget sweep and seed variation.

\begin{table*}[t]
\centering
\caption{Accuracy (\%) on four mathematics benchmarks. Our Frozen, SFT, and \method{} evaluations use a 768-token cap. CODI, MUX, and LOTUS use their source protocols, including reproductions. $^\dagger$ and $^\ddagger$ mark results quoted from MUX \citep{suleymanzade2026mux} and LOTUS \citep{fan2026lotus}, respectively. \emph{OOD} is the mean of SVAMP, GSM-Hard, and MultiArith. Shading marks \method{}, with $\pm$ denoting standard deviation over three encoder seeds. Best and second-best latent methods are \textbf{bold} and \underline{underlined}. A dash marks a configuration for which no comparable trained checkpoint is available at that scale.}
\label{tab:main_results}
\definecolor{INTACTBlue}{RGB}{235,243,252}
\small
\renewcommand{\arraystretch}{0.95}
\setlength{\tabcolsep}{5pt}
\setlength{\aboverulesep}{0.15ex}\setlength{\belowrulesep}{0.15ex}
\setlength{\abovetopsep}{0pt}\setlength{\belowbottomsep}{0pt}
\resizebox{\ifdim\width>\linewidth\linewidth\else\width\fi}{!}{%
\begin{tabular}{cl ccccc}
\toprule
\textbf{Backbone} & \textbf{Method} & \textbf{GSM8K}\,$\uparrow$ & \textbf{SVAMP}\,$\uparrow$ & \textbf{GSM-Hard}\,$\uparrow$
& \textbf{MultiArith}\,$\uparrow$ & \textbf{OOD}\,$\uparrow$ \\
\midrule
\multirow{8}{*}{\textsc{Llama-3.2-1B}}
& No-CoT$^\dagger$           & 30.9 & 44.1 & 7.1 & 70.9 & 40.7 \\
& Aug-SFT$^\dagger$          & 61.6 & 66.7 & 15.6 & 99.3 & 60.5 \\
& Nemo-SFT                   & 53.1 & 59.3 & 15.0 & 92.8 & 55.7 \\
\cmidrule(lr){2-7}
& LOTUS$^\ddagger$           & \textbf{57.3} & 60.9 & 12.7 & \underline{98.3} & \underline{57.3} \\
& MUX$^\dagger$              & \underline{56.7} & \textbf{63.6} & 13.0 & \textbf{98.5} & \textbf{58.4} \\
& CODI$^\dagger$             & 55.6 & \underline{61.1} & 12.8 & 96.1 & 56.7 \\
\rowcolor{INTACTBlue}
& \method{} Stage 1          & 44.2\,{\scriptsize$\pm$~1.0} & 55.3\,{\scriptsize$\pm$~2.3} & \underline{13.1}\,{\scriptsize$\pm$~0.6} & 85.6\,{\scriptsize$\pm$~0.1} & 51.3\,{\scriptsize$\pm$~1.0} \\
\rowcolor{INTACTBlue}
& \method{} Stage 2          & 48.7\,{\scriptsize$\pm$~1.3} & 59.8\,{\scriptsize$\pm$~1.3} & \textbf{13.7}\,{\scriptsize$\pm$~0.3} & 89.9\,{\scriptsize$\pm$~0.2} & 54.5\,{\scriptsize$\pm$~0.4} \\
\midrule
\multirow{8}{*}{\textsc{Llama-3.2-3B}}
& No-CoT$^\dagger$           & 38.3 & 52.9 & 9.5 & 88.7 & 50.4 \\
& Aug-SFT$^\dagger$          & 71.5 & 71.0 & 17.0 & 98.3 & 62.1 \\
& Nemo-SFT                   & 82.9 & 83.5 & 30.3 & 97.8 & 70.5 \\
\cmidrule(lr){2-7}
& LOTUS$^\ddagger$           & 70.0 & 75.7 & 16.0 & \underline{99.9} & 63.9 \\
& MUX$^\dagger$              & 65.0 & 77.1 & 15.2 & \textbf{100.0} & 64.1 \\
& CODI$^\dagger$             & 60.8 & 73.3 & 14.3 & 98.7 & 62.1 \\
\rowcolor{INTACTBlue}
& \method{} Stage 1          & \underline{77.9}\,{\scriptsize$\pm$~0.3} & \underline{81.1}\,{\scriptsize$\pm$~0.6} & \textbf{27.9}\,{\scriptsize$\pm$~0.2} & 98.3\,{\scriptsize$\pm$~0.5} & \textbf{69.1}\,{\scriptsize$\pm$~0.3} \\
\rowcolor{INTACTBlue}
& \method{} Stage 2          & \textbf{79.3}\,{\scriptsize$\pm$~0.4} & \textbf{81.8}\,{\scriptsize$\pm$~0.9} & \underline{27.0}\,{\scriptsize$\pm$~0.7} & 97.7\,{\scriptsize$\pm$~0.3} & \underline{68.8}\,{\scriptsize$\pm$~0.2} \\
\midrule
\multirow{8}{*}{\textsc{Llama-3.1-8B}}
& No-CoT                     & 26.5 & 57.2 & 7.5 & 79.8 & 48.2 \\
& Aug-SFT                    & 74.2 & 77.3 & 26.5 & 97.3 & 67.0 \\
& Nemo-SFT                   & 87.3 & 86.3 & 35.9 & 98.0 & 73.4 \\
\cmidrule(lr){2-7}
& LOTUS                      & -- & -- & -- & -- & -- \\
& MUX                        & 65.4 & 77.8 & 16.1 & \textbf{100.0} & 64.6 \\
& CODI                       & -- & -- & -- & -- & -- \\
\rowcolor{INTACTBlue}
& \method{} Stage 1          & \textbf{83.7}\,{\scriptsize$\pm$~0.6} & \underline{82.8}\,{\scriptsize$\pm$~1.6} & \underline{33.2}\,{\scriptsize$\pm$~0.7} & 97.4\,{\scriptsize$\pm$~1.1} & \underline{71.1}\,{\scriptsize$\pm$~1.1} \\
\rowcolor{INTACTBlue}
& \method{} Stage 2          & \underline{83.6}\,{\scriptsize$\pm$~0.7} & \textbf{84.6}\,{\scriptsize$\pm$~1.1} & \textbf{33.4}\,{\scriptsize$\pm$~1.3} & \underline{98.4}\,{\scriptsize$\pm$~0.6} & \textbf{72.1}\,{\scriptsize$\pm$~0.7} \\
\midrule
\multirow{8}{*}{\textsc{Qwen2.5-3B}}
& No-CoT                     & 27.6 & 52.5 & 8.7 & 63.7 & 41.6 \\
& Aug-SFT                    & 74.8 & 76.2 & 34.6 & 97.7 & 69.5 \\
& Nemo-SFT                   & 81.8 & 86.8 & 45.3 & 97.5 & 76.5 \\
\cmidrule(lr){2-7}
& LOTUS                      & 66.7 & 71.6 & 17.9 & \textbf{100.0} & 63.2 \\
& MUX                        & -- & -- & -- & -- & -- \\
& CODI                       & -- & -- & -- & -- & -- \\
\rowcolor{INTACTBlue}
& \method{} Stage 1          & \underline{78.0}\,{\scriptsize$\pm$~2.0} & \textbf{87.1}\,{\scriptsize$\pm$~0.4} & \underline{40.8}\,{\scriptsize$\pm$~2.2} & 97.0\,{\scriptsize$\pm$~0.6} & \underline{75.0}\,{\scriptsize$\pm$~1.0} \\
\rowcolor{INTACTBlue}
& \method{} Stage 2          & \textbf{79.7}\,{\scriptsize$\pm$~0.5} & \underline{85.5}\,{\scriptsize$\pm$~0.1} & \textbf{41.4}\,{\scriptsize$\pm$~1.1} & \underline{98.5}\,{\scriptsize$\pm$~0.6} & \textbf{75.1}\,{\scriptsize$\pm$~0.5} \\
\bottomrule
\end{tabular}
}%
\end{table*}

\paragraph{Comparison with latent methods.}
\hyperref[tab:main_results]{Table~\ref*{tab:main_results}} reports accuracy on the four backbones. At \textsc{Llama-3.2-1B}, \method{} remains below \augsft{}
and the latent baselines, reaching $48.7$ on GSM8K and $54.5$ out of domain, compared with
$61.6$ and $60.5$ for \augsft{}. This ordering
reverses at 3B, where \method{} reaches
$79.3$ and $68.8$, outperforming the strongest published latent baseline by $9.3$ and $4.7$
points and \augsft{} by $7.8$ and $6.7$ points. The advantage over \augsft{} persists at 8B
($83.6/72.1$ versus $74.2/67.0$) and on \textsc{Qwen2.5-3B} ($79.7/75.1$ versus
$74.8/69.5$), showing that it extends beyond the Llama family. The margin over the latent
baselines therefore widens with scale, from $-8.6$ on GSM8K at 1B to $+9.3$ at 3B and
$+18.2$ at 8B.
\augsft{} is strongest at 1B, whereas \nemosft{} leads on GSM8K and the out-of-domain
average at 3B and above, exceeding Stage~2 by $2.1$--$3.7$ and $1.3$--$1.7$ points,
respectively. Thus, \method{} does not uniformly outperform supervised fine-tuning.
It improves the frozen model and achieves the strongest GSM8K and average results among
the evaluated latent methods at 3B and above, without modifying the language model.

\par\WFclear

\FloatBarrier

\subsection{Inference Efficiency}
\label{sec:efficiency}

\begin{wrapfigure}{r}{0.52\textwidth}
\vspace{-1.6\baselineskip}
\centering
\includegraphics[width=\linewidth]{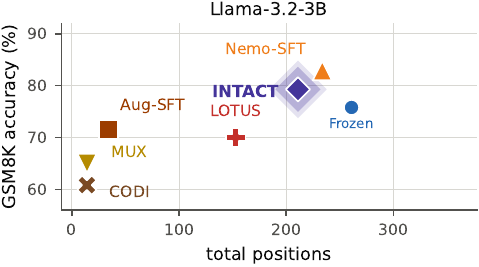}
\captionsetup{skip=0pt}
\caption{GSM8K accuracy versus total positions on \textsc{Llama-3.2-3B}, with
\method{} trained on GSM8K-Aug. Frozen, \method{}, and \nemosft{} use a 768-token cap. CODI, MUX, and LOTUS
follow their source protocols. Counts include latent positions where applicable.}
\label{fig:latent_comparison}
\vspace{-1.4\baselineskip}
\end{wrapfigure}

\paragraph{Accuracy--generation cost.}
We measure cost in total positions, defined as generated tokens plus
\method{}'s $16$ latent positions, excluding the question.
\hyperref[fig:latent_comparison]{Figure~\ref*{fig:latent_comparison}} compares operating points on
\textsc{Llama-3.2-3B}. \method{} reaches $79.3$ accuracy at $211$ positions,
compared with $75.8$ at $261$ for the frozen backbone. \augsft{}, CODI, and
MUX occupy more compressed operating points, but reach only $71.5$, $60.8$,
and $65.0$ accuracy, respectively. Unlike CODI and MUX, \method{} retains a
textual reasoning continuation, and unlike all three it improves the
accuracy--efficiency trade-off without updating the backbone.

\paragraph{Effect of the two stages.}
\hyperref[tab:main_tokens]{Table~\ref*{tab:main_tokens}} reports generated tokens across backbones. Both
stages shorten generation in every case, and which contributes more depends on
the backbone. The effect is particularly pronounced at 8B, where generation falls
from $754$ tokens for the frozen backbone to $269$ after Stage~1 and $182$
after Stage~2.
This shortening does not generally come at the expense of accuracy. Stage~2 improves
six of the eight in-domain and out-of-domain measurements reported in \hyperref[tab:budget768]{Table~\ref*{tab:budget768}}.

\paragraph{End-to-end latency.}
Token counts provide an implementation-independent measure of generation cost,
but do not directly determine wall-clock time. We therefore measure latency on
the full GSM8K test split using \textsc{Llama-3.2-3B} at batch size one on one
H100 NVL. As \hyperref[tab:latency]{Table~\ref*{tab:latency}} shows, \method{} reduces mean latency from
$4{,}180$ to $3{,}386$ ms per problem, a $19\%$ reduction that includes
latent-prefix construction. Constructing the prefix costs $127.5$ ms, while
the longer context raises measured decoding cost from $16.03$ to $16.67$ ms
per generated token. Emitting $65$ fewer tokens more than offsets these costs,
showing that the reduction in generation length translates into a measurable,
though not proportional, reduction in latency. Both systems generate under the same bare
prompt. \hyperref[tab:frozen-prompt]{Table~\ref*{tab:frozen-prompt}} reports how the frozen backbone's token cost changes
under its chat template.

%
%
\begin{table}[t]
\centering
\captionsetup{width=0.82\linewidth}
\small
\caption{Mean generated tokens at a $768$-token cap, for encoders trained on GSM8K-Aug.
Both stages shorten generation on every backbone, in domain and out, and Stage~2 is
shorter than Stage~1 in every row. \emph{OOD} is the average over SVAMP, GSM-Hard and
MultiArith.}
\label{tab:main_tokens}
\setlength{\tabcolsep}{8pt}
\renewcommand{\arraystretch}{1.2}
\begin{tabular}{@{}l cc cc cc@{}}
\toprule
& \multicolumn{2}{c}{\textbf{Frozen LM}} & \multicolumn{2}{c}{\textbf{\method{} Stage 1}}
& \multicolumn{2}{c}{\textbf{\method{} Stage 2}} \\
\cmidrule(lr){2-3}\cmidrule(lr){4-5}\cmidrule(lr){6-7}
\textbf{Backbone} & \textbf{GSM8K} $\downarrow$ & \textbf{OOD} $\downarrow$ & \textbf{GSM8K} $\downarrow$ & \textbf{OOD} $\downarrow$
& \textbf{GSM8K} $\downarrow$ & \textbf{OOD} $\downarrow$ \\
\midrule
\textsc{Llama-3.2-1B} & 252 & 210 & 241 & 202 & \textbf{215} & \textbf{171} \\
\textsc{Llama-3.2-3B} & 261 & 224 & 225 & 190 & \textbf{195} & \textbf{165} \\
\textsc{Llama-3.1-8B} & 754 & 745 & 269 & 207 & \textbf{182} & \textbf{155} \\
\textsc{Qwen2.5-3B}   & 273 & 203 & 246 & 196 & \textbf{145} & \textbf{121} \\
\bottomrule
\end{tabular}
\end{table}

\begin{table}[t]
\centering
\captionsetup{width=0.82\linewidth}
\small
\caption{Mean end-to-end latency per GSM8K problem on \textsc{Llama-3.2-3B}, measured
over the full test split at batch~1 on one H100 NVL. The saving comes from emitting
fewer tokens, not from faster decoding.}
\label{tab:latency}
\setlength{\tabcolsep}{7pt}
\renewcommand{\arraystretch}{1.2}
\begin{tabular}{@{}l rrrrr@{}}
\toprule
\textbf{System} & \textbf{Tokens} $\downarrow$ & \textbf{ms/token} & \textbf{Prefix ms} $\downarrow$
& \textbf{Generation ms} $\downarrow$ & \textbf{Total ms} $\downarrow$ \\
\midrule
Frozen backbone & $260.8$ & $16.03$ & ---     & $4{,}179.6$ & $4{,}179.6$ ($1.00\times$) \\
\rowcolor{INTACTBlue}
\method{}       & $\mathbf{195.5}$ & $16.67$ & $127.5$ & $3{,}258.1$ & $\mathbf{3{,}385.6}$ ($\mathbf{0.81\times}$) \\
\bottomrule
\end{tabular}
\end{table}

\FloatBarrier

\subsection{Ablations}
\label{sec:ablations}

Each ablation isolates one design decision and holds the rest of the recipe fixed.
The preservation and capacity ablations use \textsc{Qwen2.5-3B}, while the substitution-depth
sweep and the architecture and prefix controls of \hyperref[tab:arch-ablation]{Table~\ref*{tab:arch-ablation}} use
\textsc{Llama-3.1-8B}. All are scored under the protocol of
\hyperref[sec:setup]{Section~\ref*{sec:setup}} with a $768$-token cap.

\newsavebox{\ablationtablebox}
\begin{figure}[!ht]
\centering
\begin{lrbox}{\ablationtablebox}
\begin{minipage}{0.525\linewidth}
\centering
\centering
\footnotesize
\definecolor{INTACTBlue}{RGB}{235,243,252}
\setlength{\tabcolsep}{3.5pt}
\renewcommand{\arraystretch}{1.13}
\resizebox{\ifdim\width>\linewidth\linewidth\else\width\fi}{!}{%
\begin{tabular}[b]{@{}l ccc@{}}
\toprule
& \textbf{GSM8K}$\uparrow$ & \textbf{OOD}$\uparrow$ & \textbf{Tok.}$\downarrow$ \\
\midrule
Frozen backbone & 77.6 & 74.6 & 273 \\
\midrule
\multicolumn{4}{@{}l}{\emph{Preservation anchor $u$}} \\
None ($\beta=0$) & 42.9 & 54.5 & \textbf{80} \\
Gold chain & 74.0 & 69.1 & 240 \\
Frozen rollout & 76.2 & \textbf{76.2} & 309 \\
\rowcolor{INTACTBlue}
Student rollout & \textbf{78.8} & 75.5 & 246 \\
\midrule
\multicolumn{4}{@{}l}{\emph{Block width $c$ (at $K{=}4$)}} \\
$c=1$ \; ($4$ pos.) & 73.7 & 72.7 & \textbf{207} \\
$c=2$ \; ($8$ pos.) & 78.0 & \textbf{76.4} & 285 \\
\rowcolor{INTACTBlue}
$c=4$ \; ($16$ pos.) & \textbf{78.8} & 75.5 & 246 \\
$c=8$ \; ($32$ pos.) & 73.7 & 72.2 & 316 \\
\bottomrule
\end{tabular}}%

\end{minipage}
\end{lrbox}
\begin{minipage}[t]{0.525\linewidth}
\vspace{0pt}
\centering
\usebox{\ablationtablebox}
\end{minipage}\hfill
\begin{minipage}[t]{0.445\linewidth}
\vspace{0pt}
\centering
\includegraphics[
  width=\linewidth,
  height=\dimexpr\ht\ablationtablebox+\dp\ablationtablebox\relax,
  keepaspectratio
]{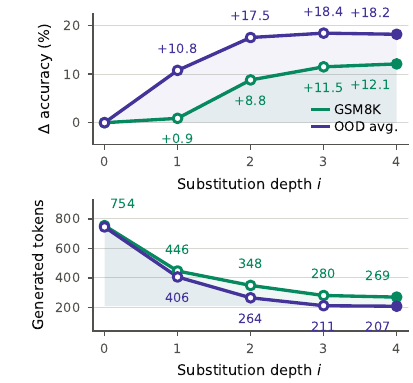}
\end{minipage}
\caption{\textbf{Left:} ablations on \textsc{Qwen2.5-3B}. Each block varies one design
decision, and shaded rows indicate settings used by \method{}. \textbf{Right:} substitution
depth sweep on \textsc{Llama-3.1-8B}. At depth $i$ only the first $i$ of $K$ blocks are
supplied, so $i=0$ is the frozen backbone and $i=K$ the full Stage~1 prefix. Both panels
use one encoder seed. \emph{OOD} is the macro-average over the three held-out benchmarks.}
\label{fig:ablations}
\end{figure}

\paragraph{Preservation anchor.}
Because $\mathcal{L}_{\mathrm{sub}}$ is scored on a reference continuation, it does
not constrain the model on its own free generations. The variants below differ only in
the continuation $\mathbf{u}$ on which $\mathcal{L}_{\mathrm{pre}}(\mathbf{u})$ is scored,
and we compare omitting the term with three choices of $\mathbf{u}$.
\begin{itemize}[leftmargin=1.6em,itemsep=0.15em,topsep=0.35em]
  \item \textbf{Gold chain.} $\mathbf{u}=[\mathbf{z}_{1:S},\mathbf{y}]$, the written trace itself, so $\mathbf{v}_{1:K}$ is
        scored while the model is given the very steps it replaces.
  \item \textbf{Frozen rollout.} $\mathbf{u}\sim p_M(\cdot\mid\mathbf{x})$, free generation from the
        backbone without the latent prefix.
  \item \textbf{Student rollout.} $\mathbf{u}\sim p_M(\cdot\mid\mathbf{x},\mathbf{v}_{1:K})$, free generation
        from the latent-conditioned model, the continuations it actually produces.
\end{itemize}

In \hyperref[fig:ablations]{Figure~\ref*{fig:ablations}} (left), omitting preservation collapses the output to $80$
tokens, so its apparent efficiency is degraded generation rather than compression. The
gold chain stabilizes training but stays below the backbone on both accuracy measures.
A frozen rollout removes that teacher forcing and gives the best out-of-domain accuracy of
the four anchors, but remains off-policy for the latent-conditioned model and lengthens
generation to $309$ tokens. Only the student rollout improves on the backbone in accuracy
and in length at once.

\paragraph{Substitution depth.}
Stage~1 supervises every available depth, so each block is learned in
the context created by its predecessors. The sweep holds the encoder fixed
and varies only the number of blocks supplied at inference. On \textsc{Llama-3.1-8B}
(\hyperref[fig:ablations]{Figure~\ref*{fig:ablations}}, right), increasing the prefix from $i=0$ to
$i=K$ raises GSM8K accuracy by $12.1$ points and the out-of-domain average by
$18.2$, while reducing generation from $754$ to $269$ tokens on GSM8K and from $745$ to
$207$ on the out-of-domain average. Accuracy therefore depends on the depth supplied, not only on
the presence of a latent prefix.

%

\paragraph{Latent capacity.}
The available substitution depth is $\min(K,S-1)$, and only $8\%$ of
GSM8K-Aug chains support five steps. We therefore fix $K=4$ and vary
the block width $c$, giving $4c$ latent positions. Performance is
non-monotonic: $c=1$ provides insufficient capacity, while $c=8$
reduces accuracy and lengthens generation. Among the narrower variants,
$c=1$ saves tokens at the cost of accuracy, whereas $c=2$ improves accuracy
but generates more tokens than the backbone. Only $c=4$ improves on the
frozen backbone while also generating fewer tokens, providing the best
accuracy--efficiency balance.

\begin{table}[t]
\centering
\caption{\textbf{Architecture and prefix controls on \textsc{Llama-3.1-8B}.}
Architecture variants retrain the encoder with the latent blocks constructed
differently. Prefix interventions instead modify a trained Stage~1
prefix only at inference. \emph{OOD} averages SVAMP, GSM-Hard and MultiArith. Each
trained arm uses one encoder seed.}
\label{tab:arch-ablation}
\small
\setlength{\tabcolsep}{10pt}
\renewcommand{\arraystretch}{1.15}
\begin{tabular}{@{}l cc cc@{}}
\toprule
& \multicolumn{2}{c}{\textbf{GSM8K}} & \multicolumn{2}{c}{\textbf{OOD}} \\
\cmidrule(lr){2-3}\cmidrule(lr){4-5}
& Acc. $\uparrow$ & Tokens $\downarrow$ & Acc. $\uparrow$ & Tokens $\downarrow$ \\
\midrule
\multicolumn{5}{@{}l}{\emph{Reference systems}} \\
Frozen backbone                                    & 72.3 & 754 & 53.9 & 745 \\
\method{} Stage~1                                  & \textbf{84.4} & 269 & 72.1 & 207 \\
\rowcolor{INTACTBlue}
\method{} Stage~2                          & 83.7 & \textbf{182} & \textbf{72.8} & \textbf{155} \\
\midrule
\multicolumn{5}{@{}l}{\emph{Architecture variants (retrained)}} \\
Independently built blocks                         & 82.4 & 767 & 69.9 & 761 \\
Jointly built blocks                               & 82.0 & 522 & 70.1 & 598 \\
\midrule
\multicolumn{5}{@{}l}{\emph{Prefix interventions (Stage~1, inference only)}} \\
Within-question block shuffle                       & 77.0 & 438 & 67.3 & 401 \\
Cross-question prefix                               & 70.9 & 513 & 64.5 & 480 \\
\bottomrule
\end{tabular}
\end{table}

\paragraph{Encoder construction.}
\hyperref[tab:arch-ablation]{Table~\ref*{tab:arch-ablation}} isolates two components of the encoder architecture.
Building the four blocks jointly, in a single encoder call, so that they can coordinate
with each other but receive no feedback from the blocks already placed, reduces GSM8K accuracy from $84.4$ to
$82.0$ and raises generation from $269$ to $522$ tokens, reaching the cap on $52.5\%$ of
problems. This variant keeps the four progressive substitution depths and differs from
its reference only in building the blocks together. \hyperref[app:train]{Appendix~\ref*{app:train}} records
what is held fixed.
Building the blocks independently, so that every block uses only the backbone's hidden
states over the question and not those induced by preceding latent blocks, leaves accuracy
at $82.4$ but increases generation from $269$ to $767$ tokens, comparable to the frozen
backbone's $754$. Both variants retain most of the accuracy of
the full construction. The clearer difference is in generation length, where the full
construction uses $269$ tokens against $767$ and $522$, so it reaches higher accuracy
while generating substantially less.

\paragraph{Prefix order and specificity.}
\hyperref[tab:arch-ablation]{Table~\ref*{tab:arch-ablation}} also tests the trained Stage~1 prefix through inference-only
interventions. Shuffling its blocks reduces GSM8K and OOD accuracy from $84.4/72.1$ to
$77.0/67.3$, while replacing it with a prefix generated for another question reduces
them to $70.9/64.5$. Both interventions also lengthen generation substantially. The prefix is therefore sensitive to both block order and the source
question, rather than acting as a generic learned prompt.

\FloatBarrier
\subsection{One Encoder Across Diverse Reasoning Domains}
\label{sec:diversedomain}

\begin{figure}[!ht]
\centering
\includegraphics[width=\linewidth]{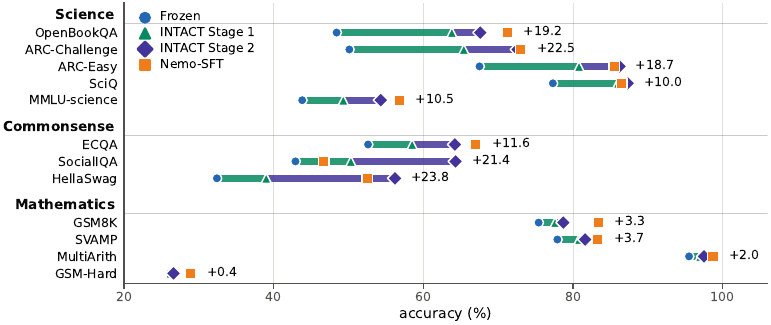}
\caption{\textbf{One encoder across twelve benchmarks.}
Accuracy on \textsc{Llama-3.2-3B} across mathematics, science, and commonsense. Each benchmark line connects the frozen backbone to Stage~2, with Stage~1 marking the intermediate result. Both \nemosft{} and \method{} are trained on the Llama-Nemotron mixture. The same \method{} encoder is used on every benchmark.}
\label{fig:domain_gains}
\end{figure}

\paragraph{Setup.}
We next test whether a single latent encoder can improve reasoning across
domains beyond GSM8K-Aug. We train on the mathematics, science, and code
solutions in Llama-Nemotron and evaluate the same encoder, without
benchmark-specific adaptation, on twelve mathematics, science, and
commonsense benchmarks. Commonsense evaluation tests transfer beyond
the domains explicitly represented in training. Every system in this
setting is scored at the $2{,}048$-token budget of \hyperref[app:eval]{Appendix~\ref*{app:eval}}, which the
long natural-language solutions of this mixture require. All systems in this setting see
the same prompt.

\paragraph{Results.}
\begin{wraptable}{r}{0.50\textwidth}
\vspace{-1.1\baselineskip}
\centering
\captionsetup{skip=0pt}
\caption{Conditioned encoder versus a learned constant-prefix control in the
diverse-domain setting on \textsc{Llama-3.2-3B}. Results use the full test
splits. \emph{Math}, \emph{Science} and \emph{Comm.} are macro-averages over
four, five and three benchmarks respectively.}
\label{tab:prefix-ablation}
\footnotesize
\setlength{\tabcolsep}{4pt}
\renewcommand{\arraystretch}{1.15}
\begin{tabularx}{\linewidth}{@{}>{\raggedright\arraybackslash}X ccc@{}}
\toprule
\textbf{System} & \textbf{Math}\,$\uparrow$ & \textbf{Science}\,$\uparrow$ & \textbf{Comm.}\,$\uparrow$ \\
\midrule
Frozen backbone & 68.8 & 57.4 & 42.6 \\
\midrule
\method{} Stage 1 & 70.4 & 69.1 & 49.3 \\
\textit{Constant prefix, Stage 1} & 70.0 & 59.9 & 44.9 \\
\midrule
\rowcolor{INTACTBlue}
\method{} Stage 2 & \textbf{71.1} & \textbf{73.6} & \textbf{61.6} \\
\textit{Constant prefix, Stage 2} & 70.9 & 64.6 & 48.5 \\
\end{tabularx}
\vspace{-1.0\baselineskip}
\end{wraptable}

\hyperref[fig:domain_gains]{Figure~\ref*{fig:domain_gains}} shows that one encoder improves the frozen
backbone on all twelve benchmarks. Gains are largest in science and
commonsense, reaching $22.5$ points on ARC-Challenge and $23.8$ on HellaSwag,
while mathematics gains range from $0.4$ to $3.7$ points. The commonsense
result is particularly notable because that domain is absent from training.
The improvements are largest where the frozen backbone is weakest, with the
exception of GSM-Hard, on which no system improves substantially.
Compared with \nemosft{}, \method{} is nearly tied on science ($1.0$ point
lower) and trails on mathematics ($2.5$ points), but leads on commonsense
($6.2$ points) in domain-average accuracy, while leaving the backbone frozen.

\paragraph{Constant-prefix control.}
To test whether the gains arise from a generic learned prompt, we replace the
encoder outputs with $Kc$ learned vectors shared across all questions and train
them with the same objective. This prefix can learn common formatting and
termination behavior but cannot encode question-specific information. \hyperref[tab:prefix-ablation]{Table~\ref*{tab:prefix-ablation}} reports it after each stage,
with the checkpoint for each selected by the same procedure as \method{}'s. The
conditioned encoder leads it on science by $9.2$ points after Stage~1 and $9.0$ after
Stage~2, and on commonsense by $4.4$ and $13.1$, while the two score similarly on
mathematics at both stages. Conditioning therefore helps where the gain over the frozen
backbone is largest, and its commonsense advantage widens with Stage~2
rather than narrowing. The mathematics tie is not evidence that
the prefix carries nothing question-specific there. Replacing a trained prefix
with one generated for a different question reduces GSM8K accuracy to $70.9$
on \textsc{Llama-3.1-8B}, below the frozen backbone's $72.3$
(\hyperref[tab:arch-ablation]{Table~\ref*{tab:arch-ablation}}).

\paragraph{Extraction and truncation.}
We test whether the gains reflect answer-extraction failures or truncation of
the frozen backbone. Answer extraction succeeds for $99.2$--$100\%$ of frozen
generations on all benchmarks except ECQA ($89.3\%$). Some generations reach
the token cap, but truncation rarely prevents scoring. HellaSwag exhibits the most
truncation, yet $99.2\%$ of its generations still yield an extractable answer. These artifacts therefore do not explain the overall gains,
although extraction failures may contribute on ECQA.

\section{Conclusion}
We studied whether latent reasoning can improve a capable language model
without updating its backbone. \method{} trains a small encoder that reads the
frozen model's own hidden states to reproduce the effect of early
written reasoning steps, first through distillation and then through latent-policy
refinement, while the language model remains frozen. Across four backbones, it
improves accuracy and reduces generation relative to the frozen model, and it
outperforms the evaluated latent-reasoning baselines on larger backbones. One
encoder trained on mathematics, science, and code also improves the same frozen
backbone on held-out mathematics, science, and commonsense benchmarks without
task-specific adaptation. Together, these results
show that latent reasoning can complement a capable model's existing reasoning,
rather than depend on rewriting its weights or maximizing compression.

\paragraph{Limitations.}
One limitation is that the latent blocks cannot be directly interpreted as
the written steps they replace. Another is that the backbone still decodes
autoregressively, so shorter generations do not necessarily produce
proportional wall-clock savings (\hyperref[sec:efficiency]{Section~\ref*{sec:efficiency}}). \method{} also
still retains language-space reasoning, so it remains more expensive than
methods that compress reasoning into a handful of tokens.

%

\bibliography{references}
\bibliographystyle{iclr2027_conference}

\appendix

%
%
\makeatletter
\providecommand{\@currentHref}{}
\def\appendixcontents@section{}
\def\appendixcontents@subsection{}
\renewcommand{\addcontentsline}[3]{%
  \@ifundefined{appendixcontents@#2}{}{%
    \addtocontents{apc}{\protect\contentsline{#2}{#3}{\thepage}{\@currentHref}}}}
\makeatother
\definecolor{AppendixBlue}{HTML}{2468B4}

\makeatletter
\renewcommand{\contentsname}{Appendix}

\renewcommand{\tableofcontents}{%
  \clearpage
  \thispagestyle{plain}%
  \vspace*{-1.2em}%
  {\fontsize{17}{21}\selectfont\sffamily\bfseries \contentsname\par}%
  \vspace{1.35em}%
  \@starttoc{apc}%
}

\renewcommand*{\l@section}[2]{%
  \ifnum\c@tocdepth>\z@
    \addpenalty\@secpenalty
    \addvspace{0.72em \@plus\p@}%
    \begingroup
      \parindent\z@
      \rightskip\@pnumwidth
      \parfillskip-\@pnumwidth
      \setlength\@tempdima{2.35em}%
      \leavevmode
      \advance\leftskip\@tempdima
      \hskip-\leftskip
      {\color{AppendixBlue}\bfseries #1}\nobreak
      \hfill\nobreak
      \hb@xt@\@pnumwidth{\hfil{\color{AppendixBlue}\bfseries #2}}\par
    \endgroup
  \fi
}

\renewcommand*{\l@subsection}[2]{%
  \ifnum\c@tocdepth>\@ne
    \begingroup
      \color{AppendixBlue}%
      \@dottedtocline{2}{1.55em}{2.85em}{#1}{\textcolor{AppendixBlue}{#2}}%
    \endgroup
  \fi
}
\makeatother

\setcounter{tocdepth}{2}
\tableofcontents
\clearpage

\section{Extended Related Work}
\label{app:related}
\hyperref[sec:related]{Section~\ref*{sec:related}} introduces the work most directly connected to \method{}. Here we
organize the broader literature by what happens to the written reasoning chain. It may remain
in language, be removed, or be represented by continuous state. This distinction matters for
\method{}, which replaces only an opening segment and leaves a frozen language model to produce
the continuation.

\subsection{From Written to Latent Reasoning}
Scratchpads and chain-of-thought make intermediate computation explicit
\citep{nye2021scratchpads,wei2022cot,kojima2022zeroshot}. More language-space computation can
improve accuracy through sampled chains \citep{wang2023selfconsistency}, test-time compute
allocation \citep{snell2024scaling,muennighoff2025s1}, process supervision
\citep{lightman2024letsverify}, or reinforcement learning on verifiable answers
\citep{deepseekai2025r1}. Its cost is autoregressive decoding, and long-reasoning models can
spend unnecessary computation on easy problems \citep{chen2024overthinking}. Prompt compression
reduces this cost while retaining a discrete, readable input \citep{jiang2023llmlingua}. After
its latent prefix, \method{} likewise returns to language-space generation.

Other methods remove or replace the written chain. Implicit chain-of-thought distills a
reasoning teacher into a student that answers directly \citep{deng2023implicit}, while stepwise
internalization progressively deletes reasoning steps from the training target
\citep{deng2024explicit}. Pause and filler tokens instead reserve computation without assigning
those positions linguistic content \citep{goyal2024pause,pfau2024filler}. Compressed
chain-of-thought predicts dense contemplation tokens as a surrogate for a trace
\citep{cheng2024ccot}, and Token Assorted combines discrete codes with text
\citep{su2025tokenassorted}. These approaches reduce or eliminate the written intermediate
sequence. \method{} is trained to re-encode its first $K$ steps and preserves a written
continuation.

Continuous reasoning admits several constructions. Coconut feeds the model's last
hidden state back as its next input embedding \citep{hao2024coconut}; CODI adds self-distillation
between explicit and latent reasoning modes \citep{shen2025codi}; and SIM-CoT supervises
individual latent steps with an auxiliary decoder \citep{wei2025simcot}. Soft Thinking is
training-free and replaces a sampled token with a probability-weighted mixture of embeddings
\citep{zhang2025softthinking}, although subsequent work questions whether this mixture provides
genuinely parallel computation \citep{wu2025singlethreaded}. Related architectures spend depth
rather than generated tokens \citep{saunshi2025looped,geiping2025recurrent}. LOTUS jointly
refines padded latent blocks through a looped backbone and supervises their token-level readout
\citep{fan2026lotus}; MUX generates latent tokens sequentially but gives each a locally decodable
target that multiplexes a span of the reference trace \citep{suleymanzade2026mux}; and KaVa uses
a compressed teacher KV-cache as its distillation target \citep{kuzina2026kava}. LaDiR instead
denoises blocks in a variational latent space whose frozen decoder reconstructs readable reasoning
steps \citep{kang2025ladir}. Surveys by \citet{zhu2025survey} and
\citet{chen2025latentsurvey} provide broader taxonomies.

\subsection{Positioning of \method{}}
\method{} is defined by two choices. The language model remains frozen, and the latents replace
written steps rather than merely adding a task-specific prompt. Prompt tuning also keeps the
model frozen, but learns one constant prefix per task \citep{lester2021prompttuning}.
SoftCoT is the closest input-dependent alternative: a small frozen assistant reads the question
and a learned projection maps its hidden states into soft tokens for a frozen backbone
\citep{xu2025softcot}. It emits those tokens in one pass and uses them as additional context.
\method{} instead derives each block from the backbone's own states, conditions it on the blocks
already placed, and trains it to replace the effect of written steps.

\hyperref[tab:positioning]{Table~\ref*{tab:positioning}} sets out the design choices that separate these methods.
\emph{Frozen} means the language model's parameters never change; \emph{reads $M$'s
states}, that the latents are computed from its hidden states rather than from its output
distribution or from the question alone; \emph{recurrent substitution}, that they are
produced in sequence, each conditioned on those already placed, and stand in for written
steps rather than supplementing them; \emph{interpretable}, that the paper recovers
token-level content from intermediate latents; and \emph{multi-domain artifact}, that one
trained artifact is evaluated across reasoning domains without per-task adaptation. The last
two columns record reported evidence, not an intrinsic limitation of methods without a checkmark.

%
%
\newcommand{\cmk}{\ensuremath{\checkmark}}
\newcommand{\xmk}{\ensuremath{\times}}
\begin{table}[!ht]
\centering
\caption{Design properties of representative latent-reasoning methods. Checkmarks
record reported design choices or evidence, not a ranking of the methods.}
\label{tab:positioning}
\small
\setlength{\tabcolsep}{5pt}
\renewcommand{\arraystretch}{1.05}
\resizebox{\ifdim\width>\linewidth\linewidth\else\width\fi}{!}{%
\begin{tabular}{@{}l ccccc@{}}
\toprule
\textbf{Method} & \textbf{Frozen} & \shortstack{\textbf{Reads $M$'s}\\\textbf{states}}
 & \shortstack{\textbf{Recurrent}\\\textbf{subst.}}
 & \textbf{Interpretable}
 & \shortstack{\textbf{Multi-domain}\\\textbf{artifact}} \\
\midrule
Coconut \citep{hao2024coconut}              & \xmk & \cmk & \cmk & \xmk & \xmk \\
CODI \citep{shen2025codi}                   & \xmk & \cmk & \cmk & \cmk & \xmk \\
SIM-CoT \citep{wei2025simcot}               & \xmk & \cmk & \cmk & \cmk & \xmk \\
LOTUS \citep{fan2026lotus}                  & \xmk & \cmk & \xmk & \cmk & \xmk \\
MUX \citep{suleymanzade2026mux}             & \xmk & \cmk & \cmk & \cmk & \xmk \\
Soft Thinking \citep{zhang2025softthinking} & \cmk & \xmk & \cmk & \cmk & \xmk \\
SoftCoT \citep{xu2025softcot}               & \cmk & \xmk & \xmk & \xmk & \xmk \\
\midrule
\textbf{\method{}}                          & \cmk & \cmk & \cmk & \xmk & \cmk \\
\bottomrule
\end{tabular}}
\end{table}

The methods that update the backbone answer a different experimental question. Coconut, CODI,
SIM-CoT, LOTUS, and MUX jointly learn a latent mechanism and alter the language model, whereas
\method{} isolates the interface by propagating gradients through fixed model weights. LOTUS also
differs structurally. Its blocks are refined jointly rather than emitted as recurrent substitutes.
CODI, SIM-CoT, LOTUS, and MUX report token-level probes or readouts of their latents. \method{}
does not make an interpretability claim.

Among the methods in \hyperref[tab:positioning]{Table~\ref*{tab:positioning}}, \method{} alone combines a frozen backbone,
backbone-state feedback, and recurrent replacement of written steps. Soft Thinking is frozen and
sequential, but constructs each soft token from an output distribution over the vocabulary rather
than from hidden states. \method{} is also the only method in the table evaluated as one trained
artifact across reasoning domains. These distinctions describe the intervention being evaluated.
They do not imply that updating the backbone or making latents directly readable is undesirable.

\paragraph{Against SoftCoT.}
SoftCoT is the closest prior learned latent interface that also keeps the language model frozen.
We use the authors' training script to train SoftCoT on
\textsc{Llama-3.1-8B-Instruct}. For a like-for-like comparison, both methods are evaluated
under the SoftCoT protocol. The \method{} row is therefore a separate single-checkpoint
evaluation and is not directly comparable to the three-seed results in
Tables~\ref{tab:budget768} and~\ref{tab:main_results}.
\hyperref[tab:softcot-comparison]{Table~\ref*{tab:softcot-comparison}} reports both accuracy and token use.

\begin{table}[H]
\centering
\caption{Comparison with SoftCoT on \textsc{Llama-3.1-8B-Instruct} under the SoftCoT
evaluation protocol. Each entry
reports accuracy (\%)/mean total tokens, including latent and generated tokens.
OOD is the macro-average over SVAMP, GSM-Hard, and MultiArith. Every entry is measured
here. We train SoftCoT with the authors' script and evaluate both methods under their
protocol. \method{} uses mathematics training (GSM8K-Aug).}
\label{tab:softcot-comparison}
\small
\setlength{\tabcolsep}{6pt}
\renewcommand{\arraystretch}{1.05}
\begin{tabular}{@{}lccccc@{}}
\toprule
\textbf{Method} & \textbf{GSM8K} & \textbf{SVAMP} & \textbf{GSM-Hard}
  & \textbf{MultiArith} & \textbf{OOD} \\
\midrule
SoftCoT & $81.0/250$ & $81.0/187$ & $\mathbf{30.6}/327$
  & $95.5/145$ & $69.0/220$ \\
\rowcolor{INTACTBlue}
\method{} & $\mathbf{82.3}/\mathbf{192}$ & $\mathbf{83.5}/\mathbf{143}$
  & $30.2/\mathbf{231}$ & $\mathbf{96.3}/\mathbf{128}$
  & $\mathbf{70.0}/\mathbf{167}$ \\
\bottomrule
\end{tabular}
\end{table}

\method{} uses fewer total tokens on every benchmark and achieves a higher OOD average,
although SoftCoT remains slightly more accurate on GSM-Hard. The separation is wider in
cost than in accuracy. \method{} leads by one point of out-of-domain accuracy and by a
$24\%$ reduction in tokens, $167$ on average compared to SoftCoT's $220$. That pattern is consistent with the two
designs, since SoftCoT's assistant reads the question alone and emits its tokens in a
single pass, whereas each \method{} block is computed from the frozen model's state after
the blocks already placed, and is trained to reproduce the effect of the written step it
replaces rather than to supply additional context.

\clearpage
\section{Notation}
\label{app:notation}
\hyperref[tab:notation]{Table~\ref*{tab:notation}} collects the symbols used in the paper.

\begin{table}[ht]
\centering
\caption{Notation used in this paper.}
\label{tab:notation}
\begin{tabular}{@{}ll@{}}
\toprule
Symbol & Meaning \\
\midrule
\multicolumn{2}{@{}l}{\emph{Model and data}}\\
$M$, $p_M$ & frozen backbone and its next-token distribution \\
$\mathcal{V}$, $d_M$ & vocabulary and input-embedding dimension of $M$ \\
$\mathbf{x}$, $\mathbf{y}$ & token sequences representing the question and reference answer \\
$\mathbf{z}_{1:S}$ & reference reasoning trace of $S$ steps, $\mathbf{z}_i\in\mathcal{V}^{n_i}$ \\
$E(\mathbf{x})$ & token embeddings of the token string $\mathbf{x}$ \\
$\hat{\mathbf{y}}$, $\operatorname{Ans}(\cdot)$ & generated reasoning-and-answer continuation and answer extraction \\
\midrule
\multicolumn{2}{@{}l}{\emph{Latent interface}}\\
$K$, $c$ & number of latent blocks and vectors per block \\
$\mathbf{v}_{1:K}$ & latent prefix, block $\mathbf{v}_i\in\mathbb{R}^{c\times d_M}$ \\
$f_\theta$, $\theta$ & latent encoder and its parameters, the only trained weights \\
$H(\cdot)$ & final-layer states of $M$ on an embedding-space input $[E(\mathbf{x}),\mathbf{v}]$ \\
$q_j$ & learned queries specific to block $j$ \\
\midrule
\multicolumn{2}{@{}l}{\emph{Stage 1, progressive substitution}}\\
$i$ & substitution depth, running to $\min(K,S-1)$ \\
$C_i^{\mathrm{T}}$, $C_i^{\mathrm{S}}$ & teacher $[\mathbf{x},\mathbf{z}_{1:i}]$ and student $[\mathbf{x},\mathbf{v}_{1:i}]$ contexts \\
$\mathbf{u}_i$ & continuation $[\mathbf{z}_{i+1:S},\mathbf{y}]$ shared by both contexts \\
$D(A\Vert B;\mathbf{u})$ & mean per-token KL between contexts $A$ and $B$ over $\mathbf{u}$ \\
$\mathcal{L}_{\mathrm{sub}}^{(i)}$ & substitution loss at depth $i$ \\
$\mathcal{L}_{\mathrm{pre}}$, $\beta$ & preservation loss on $\hat{\mathbf{y}}$ and its weight \\
\midrule
\multicolumn{2}{@{}l}{\emph{Stage 2, latent-policy refinement}}\\
$\pi_\theta$ & policy $p_M(\hat{\mathbf{y}}\mid\mathbf{x},\mathbf{v}_{1:K})$, optimized through $\theta$ only \\
$G$, $\hat{\mathbf{y}}^g$ & group size and the $g$-th sampled continuation \\
$r^g$, $A^g$ & binary correctness reward and its group-standardized advantage \\
\bottomrule
\end{tabular}

\end{table}

\section{Experimental Details}
\label{app:exp}

\subsection{\method{} Training}
\label{app:train}

\begin{figure*}[t]
    \centering
    \includegraphics[width=\textwidth]{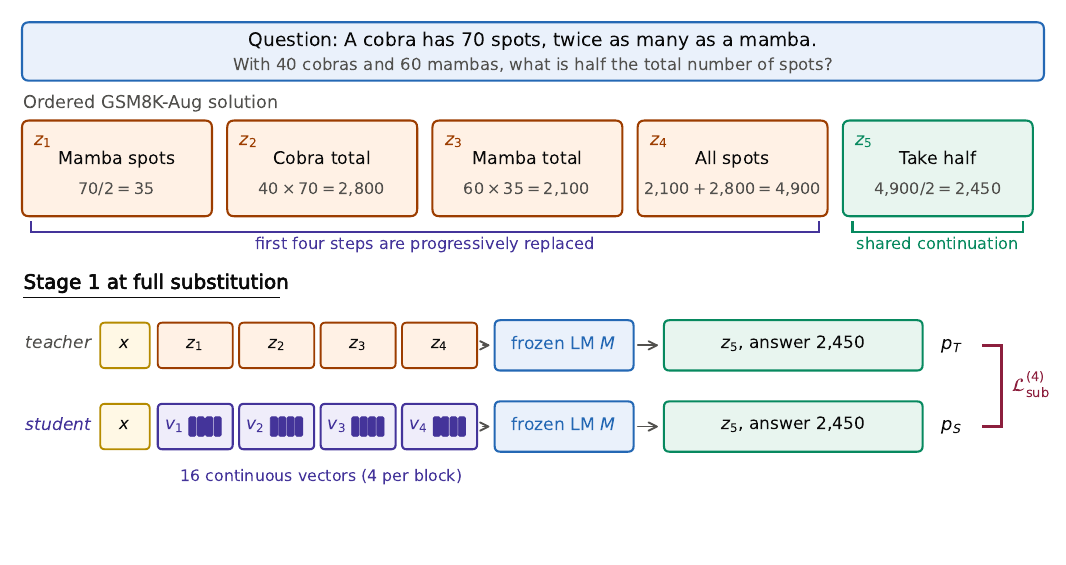}
    \caption{\textbf{Progressive substitution on a mathematics example.}
    GSM8K-Aug provides five ordered calculations. The question is shortened here for space.
    At full substitution, the student replaces the first four written steps
    $z_{1:4}$ with latent blocks $v_{1:4}$, while the final calculation $z_5$
    and answer remain the shared continuation. The frozen language model evaluates
    both prefixes, and $\mathcal{L}_{\mathrm{sub}}$ at depth $i=4$ matches the resulting
    continuation distributions. Vertical alignment denotes which written step each
    latent block replaces, not a claim that the latent vectors decode to that step.}
    \label{fig:math-example}
\end{figure*}

\paragraph{Data and depth sampling.}
For comparison against the latent-reasoning methods of
\hyperref[tab:main_results]{Table~\ref*{tab:main_results}}, we train on the corpus they use, GSM8K-Aug, described
in \hyperref[app:datasets]{Appendix~\ref*{app:datasets}}. We parse its
calculator annotations as ordered reasoning steps and retain the $322{,}712$
examples containing at least two steps. For a trace of $S$ steps, Stage~1
supervises depths $i\leq\min(K,S-1)$, leaving at least one written step in the
shared continuation. Only $55.5\%$, $23.2\%$, and $8.3\%$ of the retained
examples support depths two, three, and four, respectively. We therefore divide
the corpus into the four length buckets $S=2$, $S=3$, $S=4$, and $S\geq5$ and
sample them uniformly. \hyperref[fig:math-example]{Figure~\ref*{fig:math-example}} illustrates a complete
four-block substitution.

\paragraph{INTACT Stage 1.}
Unless otherwise stated, experiments use $K=4$ blocks of $c=4$ vectors. Each minibatch produces one
encoder update at every depth supported by its examples. Preservation rollouts
are sampled from the latent-conditioned model with temperature $0.7$ and
top-$p=0.95$, then held fixed while $\mathcal{L}_{\mathrm{pre}}$ is evaluated.
The preservation weights are $\beta=0.28$, $0.30$, $0.34$, and $0.28$ for
Llama-1B, Llama-3B, Llama-8B, and Qwen-3B, respectively.

Teacher and student prefixes can occupy different numbers of positions. Let
$\ell_z=|\mathbf{z}_{1:i}|$ and $\ell_v=ic$. We assign the latent span position
IDs beginning at
\[
  \max\!\left(|\mathbf{x}|,\,
  |\mathbf{x}|+\ell_z-\ell_v\right),
\]
which aligns the shared continuation whenever $\ell_z\geq\ell_v$ and otherwise
prevents the latent span from overlapping the question. This shift applies only
to Stage~1 substitution, where teacher and student are evaluated on the same
continuation. Preservation, Stage~2, and inference generate without a reference
trace, so the question and latent prefix use contiguous position IDs.

On GSM8K-Aug the shift is systematic. The shortest step is $5$ tokens against $c=4$
latent slots, so $\ell_z>\ell_v$ at every depth and the clamp never applies. The
displacement grows from $3$ positions at $i{=}1$ to $14$ at $i{=}4$, roughly a quarter of
the prefix, so Stage~1 trains on a layout that inference does not reproduce.

Llama encoders are trained for $1{,}500$ minibatch iterations, with checkpoints
saved every $100$ iterations. Qwen encoders are trained for $3{,}000$ iterations
and saved every $200$. We select checkpoints by generation accuracy on $300$
held-out GSM8K-Aug questions that do not overlap the GSM8K test set.

\paragraph{INTACT Stage 2.}
Each Stage~2 run begins from its selected Stage~1 checkpoint. We first evaluate
four completions for each of the first $3{,}000$ eligible training questions
and retain only questions for which the completions include both a correct and
an incorrect answer. This produces between $1{,}473$ and $1{,}906$ questions
across the Stage~1 seed lineages. During training, every visit draws eight fresh
completions. Questions whose rewards all agree contribute no update.

An optimizer step contains $12$ questions and eight completions per question,
sampled at temperature $1.0$. We optimize the per-sequence-normalized objective
for $800$ steps using AdamW with learning rate $10^{-5}$, zero weight decay,
and gradient clipping at $1.0$. The reported checkpoint is selected by
generation accuracy on the held-out GSM8K-Aug split.

The frozen language model is kept in bfloat16 and the encoder in float32.
Gradients pass through the language model, but only the encoder is updated.
Each run uses one NVIDIA H100 NVL. \hyperref[tab:hparams]{Table~\ref*{tab:hparams}} summarizes the
optimization settings. Reported uncertainty is the sample standard deviation over
the corresponding three encoder seed lineages.

%
\begin{table}[t]
\centering
\caption{Hyperparameters for both stages. The optimizer is AdamW throughout.}
\label{tab:hparams}
\footnotesize
\setlength{\tabcolsep}{6pt}
\renewcommand{\arraystretch}{1.15}
\begin{tabular}{@{}l cccc@{}}
\toprule
& \textsc{Llama-3.2} & \textsc{Llama-3.2} & \textsc{Llama-3.1} & \textsc{Qwen2.5} \\
& 1B-Instruct & 3B-Instruct & 8B-Instruct & 3B (base) \\
\midrule
\multicolumn{5}{@{}l}{\textbf{Stage 1: progressive substitution}} \\
latent blocks $K$            & $4$ & $4$ & $4$ & $4$ \\
vectors per block $c$        & $4$ & $4$ & $4$ & $4$ \\
learning rate                & $10^{-4}$ & $10^{-4}$ & $10^{-4}$ & $10^{-4}$ \\
weight decay                 & $0.01$ & $0.01$ & $0.01$ & $0.01$ \\
minibatch size               & $8$ & $8$ & $8$ & $8$ \\
gradient clipping            & $1.0$ & $1.0$ & $1.0$ & $1.0$ \\
preservation weight $\beta$  & $0.28$ & $0.30$ & $0.34$ & $0.28$ \\
\midrule
\multicolumn{5}{@{}l}{\textbf{Stage 2: latent-policy refinement}} \\
learning rate                & $10^{-5}$ & $10^{-5}$ & $10^{-5}$ & $10^{-5}$ \\
weight decay                 & $0$ & $0$ & $0$ & $0$ \\
gradient clipping            & $1.0$ & $1.0$ & $1.0$ & $1.0$ \\
rollouts per question        & $8$ & $8$ & $8$ & $8$ \\
sampling temperature         & $1.0$ & $1.0$ & $1.0$ & $1.0$ \\
\bottomrule
\end{tabular}
\end{table}

\paragraph{Architecture controls.}
The two architecture variants of \hyperref[tab:arch-ablation]{Table~\ref*{tab:arch-ablation}} are retrained rather than
modified at inference, and the jointly built variant is compared against a reference run
that differs from it in one setting alone. Both draw the same GSM8K-Aug stream, take the
same teacher continuations at the same four substitution depths, carry preservation
weight $\beta=0.34$ against the student's own rollout, train for $1{,}500$ steps at $K=4$
and so take the same number of optimizer updates, and select a checkpoint by generation
accuracy on the held-out validation split. The jointly built encoder flattens all $Kc$
queries into one encoder pass, so its blocks attend to one another and receive the
progressive losses at the same four depths. What it does not see is the backbone's state
after the preceding blocks have been placed.

\subsection{Diverse-Domain Training and Evaluation}
\label{app:domainsetup}

\paragraph{Training corpus.}
We construct a $12{,}000$-example training stream from Llama-Nemotron. Checkpoints
are selected on $200$ held-out GSM8K-Aug validation questions that are disjoint from
the GSM8K test set. The resulting mixture is $64.3\%$ mathematics,
$21.4\%$ science, and $14.3\%$ code. For mathematics, we use the
reasoning-off collection of direct solutions from
\textsc{Llama-3.3-70B-Instruct}. The reasoning-on collection instead places
the usable solution after a substantially longer exploratory trace. The
science and code examples are shared by both collections. We convert responses
to ordered steps using equation-bearing spans for mathematics, sentences for
science, and lines of fenced programs for code.

\paragraph{Training.}
We train the same $107$M encoder architecture on
\textsc{Llama-3.2-3B-Instruct}, with $K=4$ blocks of $c=4$ vectors. Stage~1
runs for $1{,}500$ steps with learning rate $10^{-4}$, batch size $8$, a
$200$-step warmup, and preservation weight $\beta=0.30$, using all three
domains. Stage~2 starts from that checkpoint and runs for $800$ group-relative policy-gradient steps with
learning rate $10^{-5}$, batch size $12$, and groups of eight rollouts sampled
at temperature $1.0$. Its rollout pool is $75\%$ Nemotron mathematics and
$25\%$ Nemotron science, because code is not used in Stage~2. Mathematics receives a
numeric-correctness reward, and science receives an option-letter correctness
reward.
Rollouts are capped at $256$ tokens, and no KL penalty to a reference policy is
used. All other settings follow \hyperref[app:train]{Appendix~\ref*{app:train}} and the language model
remains frozen throughout.

\paragraph{Evaluation.}
We evaluate under the sampling protocol of \hyperref[app:eval]{Appendix~\ref*{app:eval}} with a
$2{,}048$-token budget. The loose budget limits the censoring caused by
unfinished frozen-model generations, which still run to the cap on a larger
fraction of the mathematics examples than Stage~2 does. Reported \method{}
costs include both the generated tokens and the $16$ latent positions.

\subsection{Baselines}
\label{app:baselines}

\paragraph{Result provenance.}
At \textsc{Llama-3.2-1B} and \textsc{Llama-3.2-3B}, we quote No-CoT,
\augsft{}, MUX, and CODI from \citet{suleymanzade2026mux}, and LOTUS from
\citet{fan2026lotus}. Measured entries at \textsc{Llama-3.1-8B} and
\textsc{Qwen2.5-3B} are ours. No directly comparable latent-reasoning
checkpoint has been released for these backbones. Published 8B MUX and CODI
results use base models, whereas our 8B backbone is instruction-tuned.

\paragraph{No-CoT and \augsft{}.}
Our No-CoT and \augsft{} runs at 8B and on Qwen use the same script and differ only in the supervision
target. We apply rank-$128$ LoRA to the
\texttt{q,k,v,o,gate,up,down} projections of every layer, with
$\alpha=2r$, dropout $0.05$, and no bias. Training uses AdamW with learning
rate $10^{-4}$, zero weight decay, $100$ warmup steps followed by cosine
decay, gradient clipping at $1.0$, sequence length $512$, and effective batch
size $16$ in bfloat16. Both train on all $385{,}530$ GSM8K-Aug examples, where
\method{} uses the $322{,}712$ of them that carry at least two steps
(\hyperref[app:train]{Appendix~\ref*{app:train}}). The adapters follow the same rank-$128$ configuration at every
scale.

\paragraph{\nemosft{}.}
We train this baseline on all four backbones using the same script and LoRA
configuration as \augsft{}, but replace GSM8K-Aug with the Llama-Nemotron
mixture from \hyperref[app:domainsetup]{Appendix~\ref*{app:domainsetup}}. Thus the two baselines differ
only in their supervision. \augsft{} uses terse calculator expressions
averaging roughly $31$ tokens, and \nemosft{} uses natural-language solutions
averaging roughly $750$ tokens. We select the step-$200$ checkpoint on every
backbone.

\paragraph{Latent-reasoning baselines.}
For LOTUS on \textsc{Qwen2.5-3B}, we run the authors' code with
$c_{\mathrm{thought}}=25$, six latent loops, constant AdamW learning rate
$5\times10^{-5}$, effective batch size $128$, and seven epochs. This covers
the curriculum through \texttt{max\_latent\_stage=6}. We report the final
epoch using the authors' evaluation script. The quoted MUX results use the
shared LoRA configuration reported in its Table~8: rank $128$, $\alpha=32$,
and dropout $0.1$. The quoted CODI results come from the same codebase and
configuration, with CODI's additional hidden-size projector included in its
trainable-parameter count. We also train MUX on \textsc{Llama-3.1-8B-Instruct},
where no checkpoint is released. MUX's published learning rate of $8\times10^{-4}$ collapses
on this backbone, so we use $10^{-4}$ with gradient clipping at $1.0$ and warmup
ratio $0.10$, and otherwise follow their \texttt{train\_llama8b\_gsm8k\_aug.sh}. \hyperref[fig:params]{Figure~\ref*{fig:params}} additionally includes SIM-CoT at 3B.
We take its GSM8K and out-of-domain accuracies from Table~2 of
\citet{suleymanzade2026mux} and estimate its trainable parameters as the shared
rank-$128$ LoRA adapter plus a full auxiliary decoder with $P_{\mathrm{dec}}=P$,
following their Appendix~B.4.

\subsection{Evaluation Protocol}
\label{app:eval}

\paragraph{Prompting and decoding.}
For numerical benchmarks, the model receives the bare question followed by a
newline. Multiple-choice prompts instead use \texttt{Question:} followed by the
question and lettered options. Neither format includes few-shot demonstrations
or a model-specific chat template.
Following the CODI/MUX protocol, we sample with temperature $0.1$, top-$k$
$40$, and top-$p$ $0.95$. Quoted results retain the protocol of their source.

\paragraph{Generation budget.}
The GSM8K-Aug mathematics experiments use a $768$-token cap for every system, and
the Llama-Nemotron diverse-domain experiment uses $2{,}048$. The larger budget reflects that mixture's longer
targets. Its steps are complete natural-language solutions rather than compact
arithmetic, and at a $512$-token cap the frozen backbone is truncated on up to $13.9\%$
of mathematics problems, suppressing the reference point the gains are measured against.
\hyperref[app:budget]{Appendix~\ref*{app:budget}} sweeps the cap from $128$ to $768$ tokens. From $512$ to
$768$ tokens, most systems move by tenths of a point, \textsc{Qwen2.5-3B} still gains $1.3$,
and the frozen \textsc{Llama-3.1-8B} loses accuracy because it rarely
terminates. \hyperref[tab:softcot-comparison]{Table~\ref*{tab:softcot-comparison}} instead adopts the
$1{,}024$-token cap of the SoftCoT protocol for both of its rows. Quoted results retain
the budget of their source.

\paragraph{Answer extraction.}
We discard text following a newly generated \texttt{Question:}, which base
models sometimes emit after completing a solution. For numerical benchmarks,
we extract, in order,
a value following \texttt{\#\#\#\#} or \texttt{\#\#\#}, an explicit
``answer is'' phrase, a boxed value, or the final number in the remaining
completion. A prediction is correct when the extracted value matches the gold
answer within $10^{-4}$. For multiple-choice benchmarks, we extract an explicit
answer letter first, then the last matching option text, and finally the last
standalone letter. Correctness is exact match with the gold option letter.

\clearpage
\section{Extended Results}
\label{app:add}

\subsection{INTACT and the Frozen Backbone Across Budgets}
\label{app:budget}
\begin{table*}[htbp]
\centering
\caption{Accuracy (\%) as a function of the generation budget, with the encoder trained on GSM8K-Aug. \emph{GSM8K} is the matching test set and \emph{OOD} averages SVAMP, GSM-Hard, and MultiArith. Stage~1 is progressive substitution alone and Stage~2 adds latent-policy refinement. Columns with $\pm$ show mean $\pm$ sample standard deviation over three encoder seeds.}
\label{tab:budget}
\footnotesize
\renewcommand{\arraystretch}{0.9}
\setlength{\tabcolsep}{5pt}
\resizebox{\ifdim\width>\linewidth\linewidth\else\width\fi}{!}{%
\begin{tabular}{cl cc cc cc}
\toprule
& & \multicolumn{2}{c}{\textbf{Frozen LM}}
  & \multicolumn{2}{c}{\method{} \textbf{Stage 1}}
  & \multicolumn{2}{c}{\method{} \textbf{Stage 2}} \\
\cmidrule(lr){3-4}\cmidrule(lr){5-6}\cmidrule(lr){7-8}
\textbf{Backbone} & \textbf{Budget} & \textbf{GSM8K}\,$\uparrow$ & \textbf{OOD}\,$\uparrow$ & \textbf{GSM8K}\,$\uparrow$ & \textbf{OOD}\,$\uparrow$
& \textbf{GSM8K}\,$\uparrow$ & \textbf{OOD}\,$\uparrow$ \\
\midrule
\multirow{5}{*}{\textsc{\shortstack{Llama\\3.2-1B}}}
& 128 & 13.2 & 33.4 & 14.0\,{\scriptsize$\pm$~0.6} & 35.8\,{\scriptsize$\pm$~0.2} & 17.3\,{\scriptsize$\pm$~2.6} & 42.1\,{\scriptsize$\pm$~3.0} \\
& 256 & 37.8 & 47.3 & 40.5\,{\scriptsize$\pm$~0.9} & 50.8\,{\scriptsize$\pm$~0.9} & 45.5\,{\scriptsize$\pm$~1.2} & 54.1\,{\scriptsize$\pm$~0.4} \\
& 384 & 43.7 & 47.8 & 43.9\,{\scriptsize$\pm$~1.0} & 51.3\,{\scriptsize$\pm$~1.0} & 48.5\,{\scriptsize$\pm$~1.4} & 54.4\,{\scriptsize$\pm$~0.5} \\
& 512 & 44.0 & 47.8 & 44.2\,{\scriptsize$\pm$~1.0} & 51.3\,{\scriptsize$\pm$~0.9} & 48.8\,{\scriptsize$\pm$~1.3} & 54.5\,{\scriptsize$\pm$~0.4} \\
& 768 & 44.0 & 47.8 & 44.2\,{\scriptsize$\pm$~1.0} & 51.3\,{\scriptsize$\pm$~1.0} & 48.7\,{\scriptsize$\pm$~1.3} & 54.5\,{\scriptsize$\pm$~0.4} \\
\midrule
\multirow{5}{*}{\textsc{\shortstack{Llama\\3.2-3B}}}
& 128 & 15.3 & 40.0 & 15.2\,{\scriptsize$\pm$~0.7} & 42.6\,{\scriptsize$\pm$~1.0} & 24.7\,{\scriptsize$\pm$~1.7} & 52.6\,{\scriptsize$\pm$~0.9} \\
& 256 & 61.6 & 64.1 & 65.0\,{\scriptsize$\pm$~0.7} & 66.9\,{\scriptsize$\pm$~0.4} & 72.2\,{\scriptsize$\pm$~0.9} & 67.5\,{\scriptsize$\pm$~0.1} \\
& 384 & 74.1 & 66.2 & 76.4\,{\scriptsize$\pm$~0.5} & 68.8\,{\scriptsize$\pm$~0.3} & 78.4\,{\scriptsize$\pm$~0.3} & 68.7\,{\scriptsize$\pm$~0.3} \\
& 512 & 75.8 & 66.5 & 77.6\,{\scriptsize$\pm$~0.3} & 69.0\,{\scriptsize$\pm$~0.3} & 79.2\,{\scriptsize$\pm$~0.3} & 68.8\,{\scriptsize$\pm$~0.2} \\
& 768 & 75.8 & 66.5 & 77.9\,{\scriptsize$\pm$~0.3} & 69.1\,{\scriptsize$\pm$~0.3} & 79.3\,{\scriptsize$\pm$~0.4} & 68.8\,{\scriptsize$\pm$~0.2} \\
\midrule
\multirow{5}{*}{\textsc{\shortstack{Llama\\3.1-8B}}}
& 128 & 14.9 & 38.9 & 15.1\,{\scriptsize$\pm$~0.7} & 41.0\,{\scriptsize$\pm$~1.2} & 30.7\,{\scriptsize$\pm$~2.8} & 55.8\,{\scriptsize$\pm$~1.6} \\
& 256 & 62.8 & 63.7 & 66.6\,{\scriptsize$\pm$~0.7} & 68.2\,{\scriptsize$\pm$~0.7} & 78.9\,{\scriptsize$\pm$~2.1} & 71.3\,{\scriptsize$\pm$~0.4} \\
& 384 & 76.3 & 62.1 & 81.8\,{\scriptsize$\pm$~0.8} & 70.9\,{\scriptsize$\pm$~0.9} & 83.3\,{\scriptsize$\pm$~0.6} & 72.1\,{\scriptsize$\pm$~0.7} \\
& 512 & 75.7 & 57.0 & 83.4\,{\scriptsize$\pm$~0.6} & 71.1\,{\scriptsize$\pm$~1.0} & 83.6\,{\scriptsize$\pm$~0.6} & 72.1\,{\scriptsize$\pm$~0.7} \\
& 768 & 72.3 & 53.9 & 83.7\,{\scriptsize$\pm$~0.6} & 71.1\,{\scriptsize$\pm$~1.1} & 83.6\,{\scriptsize$\pm$~0.7} & 72.1\,{\scriptsize$\pm$~0.7} \\
\midrule
\multirow{5}{*}{\textsc{\shortstack{Qwen\\2.5-3B}}}
& 128 & 18.5 & 41.9 & 28.2\,{\scriptsize$\pm$~3.1} & 47.3\,{\scriptsize$\pm$~2.5} & 52.3\,{\scriptsize$\pm$~0.8} & 68.4\,{\scriptsize$\pm$~0.3} \\
& 256 & 51.6 & 68.1 & 57.3\,{\scriptsize$\pm$~0.8} & 69.3\,{\scriptsize$\pm$~0.2} & 78.7\,{\scriptsize$\pm$~0.5} & 74.8\,{\scriptsize$\pm$~0.5} \\
& 384 & 70.3 & 73.1 & 72.9\,{\scriptsize$\pm$~2.0} & 73.6\,{\scriptsize$\pm$~1.0} & 79.7\,{\scriptsize$\pm$~0.4} & 75.1\,{\scriptsize$\pm$~0.5} \\
& 512 & 76.3 & 74.3 & 77.2\,{\scriptsize$\pm$~2.1} & 74.7\,{\scriptsize$\pm$~1.0} & 79.7\,{\scriptsize$\pm$~0.5} & 75.1\,{\scriptsize$\pm$~0.5} \\
& 768 & 77.6 & 74.6 & 78.0\,{\scriptsize$\pm$~2.0} & 75.0\,{\scriptsize$\pm$~1.0} & 79.7\,{\scriptsize$\pm$~0.5} & 75.1\,{\scriptsize$\pm$~0.5} \\
\bottomrule
\end{tabular}
}%
\vspace{0.6\baselineskip}
\end{table*}

\hyperref[tab:budget]{Table~\ref*{tab:budget}} compares the frozen backbone with both \method{} stages at every
backbone and every generation budget, of which \hyperref[tab:budget768]{Table~\ref*{tab:budget768}} shows one cap.
All budget values are scored from prefixes of the same generation pass. The frozen
\textsc{Llama-3.1-8B} row falls as the budget grows because that model rarely terminates,
reaching the cap on $97.1\%$ of GSM8K generations against $8.6\%$ for Stage~1 and none for
Stage~2. 

\paragraph{Prompt format.}
To isolate the effect of the prompt format, we re-evaluate the three instruction-tuned
Llama backbones with their own chat templates, a sample of which is shown in
\hyperref[fig:prompt-example]{Figure~\ref*{fig:prompt-example}}. The template helps at 8B, where it largely removes the
termination failure and raises GSM8K accuracy from $72.3$ to $78.4$, but the effect is
inconsistent across scales. Accuracy rises at 1B and falls at 3B. \method{} Stage~2, still scored
under the bare prompt, remains more accurate than the chat-prompted frozen model on GSM8K
and out of domain at all three scales, while its token cost is closer, lower at 8B and
higher at 1B and 3B (\hyperref[tab:frozen-prompt]{Table~\ref*{tab:frozen-prompt}}).

\providecommand{\spc}[1]{<|#1|>}
\begin{figure}[H]
\centering
\begin{minipage}{0.98\linewidth}
\footnotesize
\textbf{Bare question}\\[2pt]
\texttt{John runs 60 miles in 3 days.}\\
\texttt{How many miles per day?{\char`\\}n}

\smallskip\hrule\smallskip
\textbf{Llama-3.1-8B chat template}\\[2pt]
\texttt{\spc{begin\_of\_text}}\\
\texttt{\spc{start\_header\_id}system\spc{end\_header\_id}}\\
\texttt{Cutting Knowledge Date: December 2023}\\
\texttt{Today Date: 26 Jul 2024}\\
\texttt{\spc{eot\_id}}\\
\texttt{\spc{start\_header\_id}user\spc{end\_header\_id}}\\
\texttt{John runs 60 miles in 3 days.}\\
\texttt{How many miles per day?\spc{eot\_id}}\\
\texttt{\spc{start\_header\_id}assistant\spc{end\_header\_id}}
\end{minipage}
\caption{Inputs compared for the frozen \textsc{Llama-3.1-8B-Instruct}. The
question is illustrative. The chat template adds its default system metadata
and turn markers.}
\label{fig:prompt-example}
\end{figure}

\begin{table}[ht]
\centering
\caption{Prompt sensitivity of the frozen Llama backbones at a $768$-token
cap. The two frozen rows differ only by prompt format. The \method{}
rows are Stage~2, evaluated with the bare prompt for reference. }
\label{tab:frozen-prompt}
\small
\setlength{\tabcolsep}{6pt}
\renewcommand{\arraystretch}{1.1}
\begin{tabular}{@{}ll ccc@{}}
\toprule
\textbf{Backbone} & \textbf{Condition} & \textbf{GSM8K}\,$\uparrow$ & \textbf{OOD}\,$\uparrow$ & \textbf{Tok.}\,$\downarrow$ \\
\midrule
\multirow{3}{*}{\textsc{Llama-3.2-1B}}
  & Frozen (bare) & 44.0 & 47.8 & 252 \\
  & Frozen (chat) & 47.3 & 52.2 & \textbf{165} \\
  & \cellcolor{INTACTBlue}\method{} (bare) & \cellcolor{INTACTBlue}\textbf{48.7} & \cellcolor{INTACTBlue}\textbf{54.5} & \cellcolor{INTACTBlue}215 \\
\midrule
\multirow{3}{*}{\textsc{Llama-3.2-3B}}
  & Frozen (bare) & 75.8 & 66.5 & 261 \\
  & Frozen (chat) & 71.6 & 66.1 & \textbf{172} \\
  & \cellcolor{INTACTBlue}\method{} (bare) & \cellcolor{INTACTBlue}\textbf{79.3} & \cellcolor{INTACTBlue}\textbf{68.8} & \cellcolor{INTACTBlue}195 \\
\midrule
\multirow{3}{*}{\textsc{Llama-3.1-8B}}
  & Frozen (bare) & 72.3 & 53.9 & 754 \\
  & Frozen (chat) & 78.4 & 69.6 & 186 \\
  & \cellcolor{INTACTBlue}\method{} (bare) & \cellcolor{INTACTBlue}\textbf{83.6} & \cellcolor{INTACTBlue}\textbf{72.1} & \cellcolor{INTACTBlue}\textbf{182} \\
\bottomrule
\end{tabular}
\end{table}

\subsection{Accuracy--Efficiency Operating Points}
\label{app:curves}
Figures~\ref{fig:frontier_3b}--\ref{fig:frontier_qwen} compare accuracy with total
positions. Upper left is better, and \method{} counts include all $16$ latent positions.
Figures~\ref{fig:curves_3b}--\ref{fig:curves_qwen} instead sweep
the generation budget, with dashed curves updating the language model and solid curves
keeping it frozen. \textsc{Llama-3.2-1B} is omitted because it was not evaluated on the finer
budget grid these panels require. At 8B, no comparable CODI or LOTUS checkpoint is available and the
MUX point is our reproduction (\hyperref[app:baselines]{Appendix~\ref*{app:baselines}}).
%
%
\begingroup
\setlength{\intextsep}{13pt plus 3pt minus 2pt}
\setlength{\abovecaptionskip}{3pt}
\setlength{\belowcaptionskip}{0pt}

\begin{figure}[H]
\centering
\includegraphics[width=0.85\linewidth]{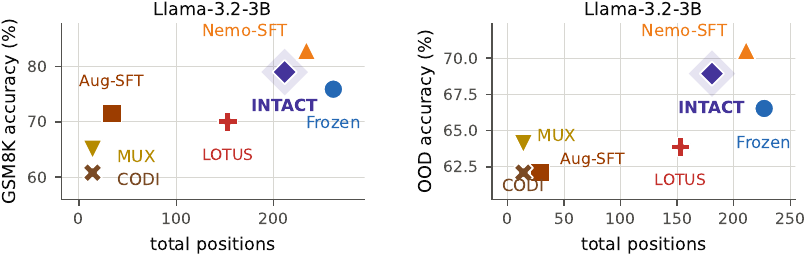}
\caption{\textsc{Llama-3.2-3B}: accuracy against total positions on GSM8K (left) and the
out-of-domain macro-average (right). Upper left is better. Frozen and \nemosft{} are
evaluated at the $768$-token cap, and every other system at the operating point it reports.}
\label{fig:frontier_3b}
\end{figure}

\begin{figure}[H]
\centering
\includegraphics[width=0.85\linewidth]{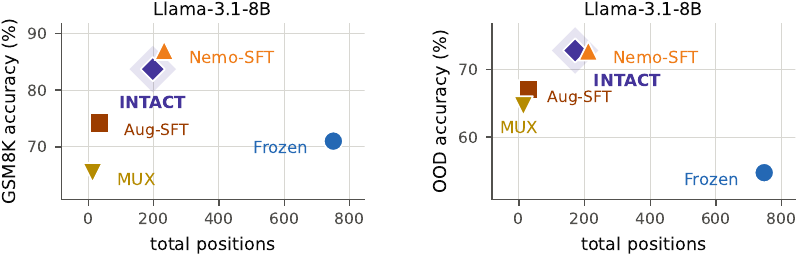}
\caption{\textsc{Llama-3.1-8B}: accuracy against total positions on GSM8K (left) and the
out-of-domain macro-average (right). Upper left is better. Frozen and \nemosft{} are
evaluated at the $768$-token cap, and every other system at the operating point it reports.}
\label{fig:frontier_8b}
\end{figure}

\begin{figure}[H]
\centering
\includegraphics[width=0.85\linewidth]{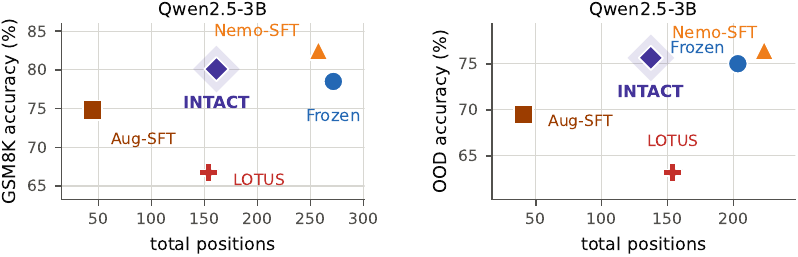}
\caption{\textsc{Qwen2.5-3B}: accuracy against total positions on GSM8K (left) and the
out-of-domain macro-average (right). Upper left is better. Frozen and \nemosft{} are
evaluated at the $768$-token cap, and every other system at the operating point it reports.}
\label{fig:frontier_qwen}
\end{figure}

\begin{figure}[H]
\centering
\includegraphics[width=0.45\linewidth]{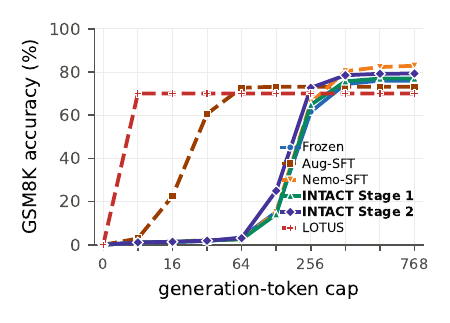}%
\hspace{0.01\linewidth}%
\includegraphics[width=0.45\linewidth]{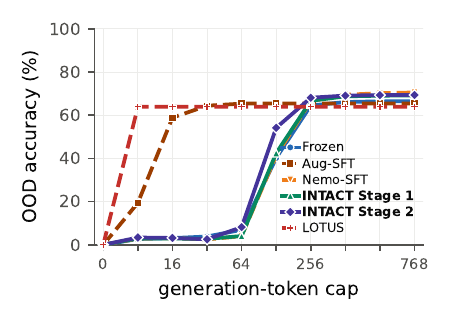}
\caption{\textsc{Llama-3.2-3B}: accuracy against the generation budget on GSM8K (left) and the
out-of-domain macro-average (right). Solid curves keep the language model frozen and dashed
curves update it. Every budget is scored from prefixes of one generation pass, so a curve
shows what that same run would have returned under a tighter cap.}
\label{fig:curves_3b}
\end{figure}

\begin{figure}[H]
\centering
\includegraphics[width=0.45\linewidth]{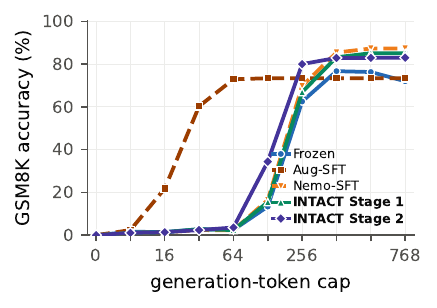}%
\hspace{0.01\linewidth}%
\includegraphics[width=0.45\linewidth]{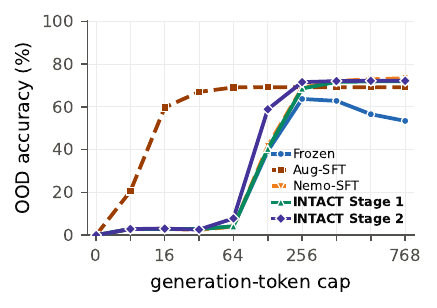}
\caption{\textsc{Llama-3.1-8B}: accuracy against the generation budget on GSM8K (left) and the
out-of-domain macro-average (right). Solid curves keep the language model frozen and dashed
curves update it. Every budget is scored from prefixes of one generation pass, so a curve
shows what that same run would have returned under a tighter cap.}
\label{fig:curves_8b}
\end{figure}

\begin{figure}[H]
\centering
\includegraphics[width=0.45\linewidth]{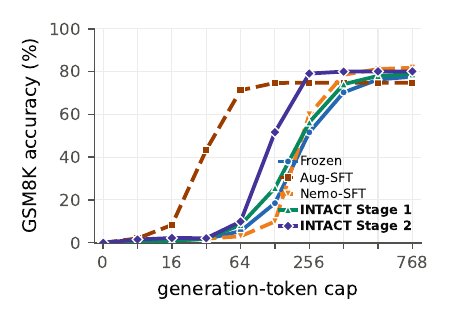}%
\hspace{0.01\linewidth}%
\includegraphics[width=0.45\linewidth]{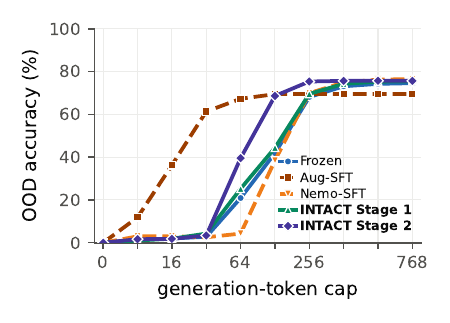}
\caption{\textsc{Qwen2.5-3B}: accuracy against the generation budget on GSM8K (left) and the
out-of-domain macro-average (right). Solid curves keep the language model frozen and dashed
curves update it. Every budget is scored from prefixes of one generation pass, so a curve
shows what that same run would have returned under a tighter cap.}
\label{fig:curves_qwen}
\end{figure}

\endgroup

\subsection{Efficiency}
\label{app:efficiency}

Three quantities are reported, and they are not interchangeable.
\begin{itemize}[leftmargin=1.6em,itemsep=0.15em,topsep=0.35em]
  \item \textbf{Generated tokens.} The text the backbone emits, excluding the latent
        prefix. \hyperref[tab:main_tokens]{Table~\ref*{tab:main_tokens}} and the ablations report this.
  \item \textbf{Total positions.} Generated tokens plus \method{}'s $Kc=16$ latent
        slots, excluding the question. The operating-point figures and
        \hyperref[tab:domain_tokens]{Table~\ref*{tab:domain_tokens}} use this.
  \item \textbf{End-to-end latency.} Wall-clock milliseconds per problem, including
        both latent-prefix construction and text generation. \hyperref[tab:latency]{Table~\ref*{tab:latency}} reports this.
\end{itemize}

\begin{figure}[H]
\centering
\includegraphics[width=0.85\linewidth]{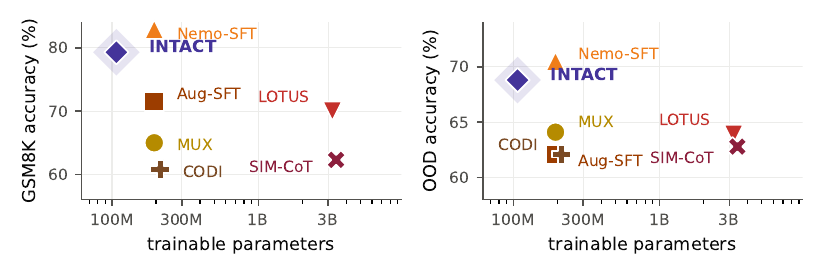}%
\caption[Parameter efficiency]{\textbf{Accuracy against trainable parameters on
\textsc{Llama-3.2-3B}}, where up and to the left is better. \emph{OOD} averages SVAMP, GSM-Hard
and MultiArith. \method{} trains a $107$M-parameter encoder and never updates the
backbone, while the baselines train between $195$M and $3.40$B. No method here reports a trainable
count, so each is derived (\hyperref[app:baselines]{Appendix~\ref*{app:baselines}}). Mathematics training (GSM8K-Aug), except \nemosft{} (Llama-Nemotron).}
\label{fig:params}
\end{figure}

\paragraph{Trainable parameters and training cost.}
\label{app:params}
\hyperref[fig:params]{Figure~\ref*{fig:params}} places the \textsc{Llama-3.2-3B} systems of
\hyperref[tab:main_results]{Table~\ref*{tab:main_results}} against the number of parameters each trains, together with
SIM-CoT. \method{} trains a $107$M encoder and never updates the backbone, so it is the
leftmost point in both panels and the most accurate of the methods that leave the backbone
frozen. The baselines run from $194.5$M for the LoRA systems to $3.40$B for SIM-CoT, which
trains an adapter alongside a full auxiliary decoder and so updates more than a full
fine-tune does. \nemosft{} is more accurate, but reaches that accuracy by updating $194.5$M
backbone parameters. No paper in this comparison reports its own trainable count, so each
is derived from the method's description. \hyperref[app:baselines]{Appendix~\ref*{app:baselines}} gives the sources
and derivations.

%

Freezing the backbone also removes its parameter gradients and optimizer state, although
its activations remain necessary for gradients to reach the encoder. On one H100 NVL and \textsc{Llama-3.2-3B},
Stage~1 takes $6.2$ hours and Stage~2 a further $10.3$ hours, placing the complete
two-stage recipe below $17$ GPU-hours.

\subsection{Diverse-Domain Generalization}
\label{app:domainresults}

\paragraph{Accuracy by stage.}
\label{app:domainstages}
%
%
%
%
%
\begin{table}[H]
\centering
\small
\caption{Accuracy (\%) of one \textsc{Llama-3.2-3B} encoder across twelve benchmarks at a
$2{,}048$-token budget, trained on the Llama-Nemotron mixture rather than GSM8K-Aug. Each
system is a single checkpoint used in every domain, not a per-domain best. $\Delta$ is Stage~2 against the frozen
backbone. Every split is scored in full. Bold is the best of
the three trained systems.}
\label{tab:nemotron_stages}
\begin{tabular}{@{}llrrrrrr@{}}
\toprule
\textbf{Domain} & \textbf{Benchmark} & $\mathbf{n}$ & \textbf{Frozen} & \textbf{Stage 1}
& \textbf{Stage 2} & \textbf{Nemo-SFT} & $\mathbf{\Delta}$ \\
\midrule
\multirow{5}{*}{Science}
 & OpenBookQA            &   500 &            48.4 &            63.8 &            67.6 &   \textbf{71.2} & $+19.2$ \\
 & ARC-Challenge         &  1172 &            50.1 &            65.4 &            72.6 &   \textbf{73.0} & $+22.5$ \\
 & ARC-Easy              &  2376 &            67.5 &            80.8 &   \textbf{86.2} &            85.5 & $+18.7$ \\
 & SciQ                  &  1000 &            77.3 &            86.0 &   \textbf{87.3} &            86.5 & $+10.0$ \\
 & MMLU-science          &  1397 &            43.8 &            49.3 &            54.3 &   \textbf{56.8} & $+10.5$ \\
\midrule
\multirow{3}{*}{Commonsense}
 & ECQA                  &  2162 &            52.6 &            58.5 &            64.2 &   \textbf{67.0} & $+11.6$ \\
 & SocialIQA             &  1954 &            42.9 &            50.3 &   \textbf{64.3} &            46.7 & $+21.4$ \\
 & HellaSwag             & 10042 &            32.4 &            39.0 &   \textbf{56.2} &            52.5 & $+23.8$ \\
\midrule
\multirow{4}{*}{Mathematics}
 & GSM8K                 &  1319 &            75.4 &            77.6 &            78.7 &   \textbf{83.4} & $+3.3$ \\
 & SVAMP                 &  1000 &            77.9 &            80.8 &            81.6 &   \textbf{83.3} & $+3.7$ \\
 & MultiArith            &   600 &            95.5 &            97.0 &            97.5 &   \textbf{98.7} & $+2.0$ \\
 & GSM-Hard              &  1319 &            26.2 &            26.2 &            26.6 &   \textbf{28.9} & $+0.4$ \\
\bottomrule
\end{tabular}
\end{table}

\hyperref[tab:nemotron_stages]{Table~\ref*{tab:nemotron_stages}} gives the accuracies behind
\hyperref[fig:domain_gains]{Figure~\ref*{fig:domain_gains}}. The two stages divide the work differently by domain. In
science, Stage~1 contributes $15.4$ of the final $19.2$-point gain on OpenBookQA and
$15.3$ of $22.5$ points on ARC-Challenge. Stage~2 has its largest effect in commonsense.
It adds $14.0$ points on SocialIQA beyond the $7.4$-point Stage~1 gain. Its contribution
on mathematics is smaller but consistently positive, ranging from $0.4$ to $1.1$ points.
Across the three commonsense benchmarks, Stage~2 reaches $61.6\%$ compared with
$55.4\%$ for \nemosft{}.

\paragraph{Token cost and termination.}
\label{app:domaintokens}

%
%
%
%
\begin{table}[H]
\centering
\small
\caption{Mean total positions per problem for the encoder trained on the Llama-Nemotron
mixture, at a $2{,}048$-token budget. \method{} columns include the $16$ latent prefix positions as well
as the tokens generated, so a number here is comparable across rows and to
\hyperref[fig:hero]{Figure~\ref*{fig:hero}}. Lower is better, and bold is the cheapest system on a row.}
\label{tab:domain_tokens}
\begin{tabular}{@{}llrrrrr@{}}
\toprule
\textbf{Domain} & \textbf{Benchmark} & $\mathbf{n}$ & \textbf{Frozen} & \textbf{Stage 1}
& \textbf{Stage 2} & \textbf{Nemo-SFT} \\
\midrule
\multirow{6}{*}{Science}
 & OpenBookQA    &   500 &            277.0 &            244.0 &    \textbf{80.0} &            147.0 \\
 & ARC-Challenge &  1172 &            234.0 &            240.0 &    \textbf{82.0} &            166.0 \\
 & ARC-Easy      &  2376 &            262.0 &            237.0 &    \textbf{98.0} &            153.0 \\
 & SciQ          &  1000 &            338.0 &            191.0 &   \textbf{123.0} &            145.0 \\
 & MMLU-science  &  1397 &            376.0 &            278.0 &   \textbf{148.0} &            246.0 \\
 \cmidrule(l){2-7}
 & Mean          &       &            297.4 &            238.0 &   \textbf{106.2} &            171.4 \\
\midrule
\multirow{4}{*}{Commonsense}
 & ECQA          &  2162 &            214.0 &            146.0 &    \textbf{61.0} &            130.0 \\
 & SocialIQA     &  1954 &            177.0 &            256.0 &            143.0 &   \textbf{141.0} \\
 & HellaSwag     & 10042 &            776.0 &            756.0 &    \textbf{60.0} &            138.0 \\
 \cmidrule(l){2-7}
 & Mean          &       &            389.0 &            386.0 &    \textbf{88.0} &            136.3 \\
\midrule
\multirow{5}{*}{Mathematics}
 & GSM8K         &  1319 &            358.0 &            259.3 &            254.8 &   \textbf{241.9} \\
 & SVAMP         &  1000 &            289.2 &            201.7 &            197.8 &   \textbf{197.4} \\
 & MultiArith    &   600 &            215.4 &            176.5 &            154.7 &   \textbf{133.2} \\
 & GSM-Hard      &  1319 &            448.1 &            388.7 &   \textbf{352.1} &            357.0 \\
 \cmidrule(l){2-7}
 & Mean          &       &            327.7 &            256.6 &            239.8 &   \textbf{232.4} \\
\bottomrule
\end{tabular}
\end{table}

\hyperref[tab:domain_tokens]{Table~\ref*{tab:domain_tokens}} gives what each system spends to reach the accuracies of
\hyperref[tab:nemotron_stages]{Table~\ref*{tab:nemotron_stages}}. Stage~2 uses fewer total positions than the frozen
backbone on all twelve benchmarks while improving accuracy on each. Stage~1 is cheaper
on ten of twelve but is more expensive on ARC-Challenge and SocialIQA. The
shortening is therefore not imposed mechanically by the latent prefix. It emerges most
consistently after latent-policy refinement.

The $2{,}048$-token budget reduces but does not eliminate truncation of frozen-model
generations. Their reported means consequently remain more sensitive to the evaluation
cap than those of \method{} and \nemosft{}, which terminate more consistently.

\section{Datasets and Evaluation Splits}
\label{app:datasets}

\subsection{Training Corpora}

The experiments use two training corpora, each for a separate encoder.

\paragraph{GSM8K-Aug.}
The main mathematics experiments of \hyperref[sec:setup]{Section~\ref*{sec:setup}} use
\textbf{GSM8K-Aug} \citep{deng2023implicit}, a $385$K-example augmentation of GSM8K
whose solutions consist of ordered, compact arithmetic steps. This structure directly
supports progressive substitution. Latent block $i$ replaces written step $i$ without
additional segmentation.

\paragraph{Llama-Nemotron.}
The diverse-domain experiment of \hyperref[sec:diversedomain]{Section~\ref*{sec:diversedomain}} uses
\textbf{Llama-Nemotron post-training data} \citep{bercovich2025llamanemotron}, which
combines mathematics, science, and code under one collection process. We use its
reasoning-off subset, whose responses are direct written solutions rather than search
traces. \hyperref[app:domainsetup]{Appendix~\ref*{app:domainsetup}} describes the sampling and step extraction.

\subsection{Evaluation Benchmarks and Splits}

\begin{table}[H]
\centering
\small
\caption{Evaluation benchmarks and scored split sizes, grouped by domain. Every row uses
the complete evaluation split after ECQA de-duplication. \emph{Form} is either a numeric
answer or an $m$-way letter choice.}
\label{tab:datasets}
\begin{tabular}{@{}llrc@{}}
\toprule
\textbf{Domain} & \textbf{Benchmark} & $\mathbf{n}$ & \textbf{Form} \\
\midrule
\multirow{4}{*}{Mathematics}
 & GSM8K \citep{cobbe2021gsm8k}              &  1319 & number \\
 & GSM-Hard \citep{gao2023pal}               &  1319 & number \\
 & SVAMP \citep{patel2021svamp}              &  1000 & number \\
 & MultiArith \citep{roy2015multiarith}      &   600 & number \\
\midrule
\multirow{5}{*}{Science}
 & OpenBookQA \citep{mihaylov2018openbookqa} &   500 & 4-way \\
 & ARC-Easy \citep{clark2018arc}             &  2376 & 4-way \\
 & ARC-Challenge \citep{clark2018arc}        &  1172 & 4-way \\
 & SciQ \citep{welbl2017sciq}                &  1000 & 4-way \\
 & MMLU-science \citep{hendrycks2021mmlu}    &  1397 & 4-way \\
\midrule
\multirow{3}{*}{Commonsense}
 & ECQA \citep{aggarwal2021ecqa}             &  2162 & 5-way \\
 & SocialIQA \citep{sap2019socialiqa}        &  1954 & 3-way \\
 & HellaSwag \citep{zellers2019hellaswag}    & $10{,}042$ & 4-way \\
\bottomrule
\end{tabular}
\end{table}

\paragraph{Benchmark coverage.}
\hyperref[tab:datasets]{Table~\ref*{tab:datasets}} lists every benchmark scored in the paper. Each benchmark uses
its complete evaluation split, subject only to the ECQA de-duplication described below.
GSM8K is the matching test set for GSM8K-Aug. For the diverse-domain experiment, we
check for overlap after case-folding the question text and removing punctuation and
multiple-choice options. We find no exact question-text
match between the examples used in either training stage and any of the twelve evaluation
splits. No evaluation question is used for checkpoint selection. Selection uses the disjoint
200-question GSM8K-Aug validation split described in
\hyperref[app:domainsetup]{Appendix~\ref*{app:domainsetup}}. Multiple-choice benchmarks
share the prompt format specified in \hyperref[app:eval]{Appendix~\ref*{app:eval}}.

\paragraph{Dataset-specific conventions.}
\emph{MMLU-science} is not a released split. It aggregates the eight MMLU subjects covered
by Llama-Nemotron's science data: high-school and college physics, chemistry, and biology,
together with conceptual physics and astronomy. ECQA contains $32$ test questions that
also occur in its training split, inherited from CommonsenseQA. We remove these questions
before scoring, leaving $2{,}162$ examples.

\end{document}